\documentclass[10pt,twocolumn,letterpaper]{article}

\usepackage{cvpr}      

\newcommand{\datasetname}{VoDaSuRe\xspace}
\usepackage[inkscapelatex=false]{svg} 

\usepackage{tikz} 
\usepackage[dvipsnames]{xcolor}
\usepackage{subcaption} 
\usepackage{nicematrix} 
\usepackage[hang,flushmargin]{footmisc}

\usepackage{pifont}
\newcommand{\cmark}{\ding{51}}%
\newcommand{\xmark}{\ding{55}}%

\usepackage{comment}
\usepackage{amsfonts}

\definecolor{cvprblue}{rgb}{0.21,0.49,0.74}
\usepackage[pagebackref,breaklinks,colorlinks,allcolors=cvprblue]{hyperref}

\def\paperID{} 
\def\confName{CVPR}
\def\confYear{2026}

\title{\datasetname: A Large-Scale Dataset Revealing Domain Shift in Volumetric Super-Resolution} 
\author{
August Leander Høeg \quad Sophia Wiinberg Bardenfleth \quad Hans Martin Kjer \\ \quad Tim Bjørn Dyrby \quad Vedrana Andersen Dahl \quad Anders Bjorholm Dahl \\
Technical University of Denmark, Kgs. Lyngby, Denmark\\
{\tt\small \{aulho, soeba, hmkj, tbdy, vand, abda\}@dtu.dk}
}

\begin{document}
\maketitle
\begin{abstract}

\noindent
Recent advances in volumetric super-resolution (SR) have demonstrated strong performance in medical and scientific imaging, with transformer- and CNN-based approaches achieving impressive results even at extreme scaling factors. In this work, we show that much of this performance stems from training on downsampled data rather than real low-resolution scans. This reliance on downsampling is partly driven by the scarcity of paired high- and low-resolution 3D datasets. To address this, we introduce \datasetname, a large-scale volumetric dataset containing paired high- and low-resolution scans. When training models on \datasetname, we reveal a significant discrepancy: SR models trained on downsampled data produce substantially sharper predictions than those trained on real low-resolution scans, which smooth fine structures. Conversely, applying models trained on downsampled data to real scans preserves more structure but is inaccurate. Our findings suggest that current SR methods are overstated -- when applied to real data, they do not recover structures lost in low-resolution scans and instead predict a smoothed average. We argue that progress in deep learning-based volumetric SR requires datasets with paired real scans of high complexity, such as \datasetname. Our dataset and code are publicly available through: \url{https://augusthoeg.github.io/VoDaSuRe/}

\end{abstract}

\section{Introduction}
\label{sec:intro}

Volumetric super-resolution promises to reveal details in low-resolution 3D scans, but are today’s deep learning models truly reconstructing missing high-resolution details, or merely learning to reverse downsampling?

Recent advances in deep learning-based volumetric super-resolution (SR) have achieved impressive results. However, we show that much of this success is a consequence of the training setup rather than genuine predictive capability. 
The vast majority of volumetric SR approaches generate paired high- and low-resolution (LR) volumes by simulating degradation by downsampling the high-resolution (HR) scans \cite{Sanchez_SRGAN3D, Chen_mDCSRN_A, Chen_mDCSRN_B, Chen_mDCSRN_C, Li_MFER, Forigua_SuperFormer, Lu_novel, Wang_EDDSR, Wang_MRGD, Wang_W2AMSN}. This setup yields near-perfect reconstructions due to the overidealistic correspondence between low and high resolution data enforced by the degradation model, but severely misrepresents the discrepancies observed in actual LR data. Furthermore, volumetric benchmark datasets are dominated by medical imaging, which often lack fine structural variations, making SR tasks using these datasets largely trivial.

\begin{figure}[t]
   \centering
   \def\svgwidth{\linewidth}
    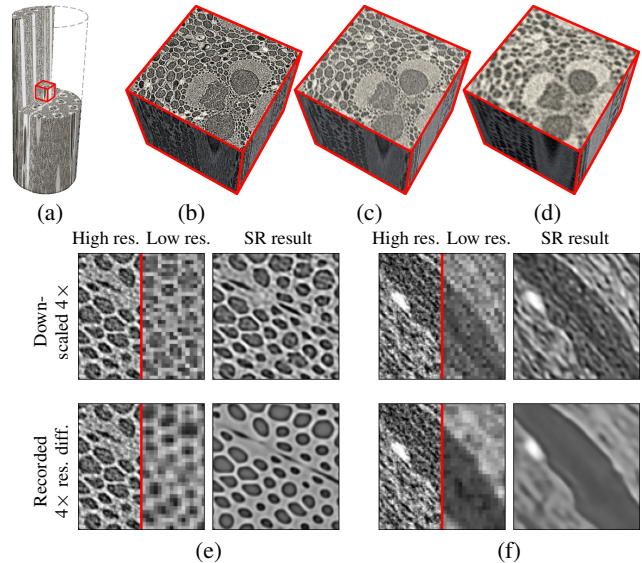
   \caption{Example from \datasetname: (a) volume of bamboo showing the cropped area in red, (b) high resolution crop, (c) 4$\times$ downsampled, (d) scan at 4$\times$ lower resolution, (e) SR of bamboo, (f) SR of cardboard. The top row in (e) and (f) is trained on downsampled HR-LR pairs, the bottom row is trained on actual LR data.}
   \label{fig:intro_overview_vodasure}
\end{figure}

\begin{figure*}[t]
    \centering
    \begin{subfigure}[b]{0.09\textwidth}
        \centering
        \includegraphics[width=\textwidth]{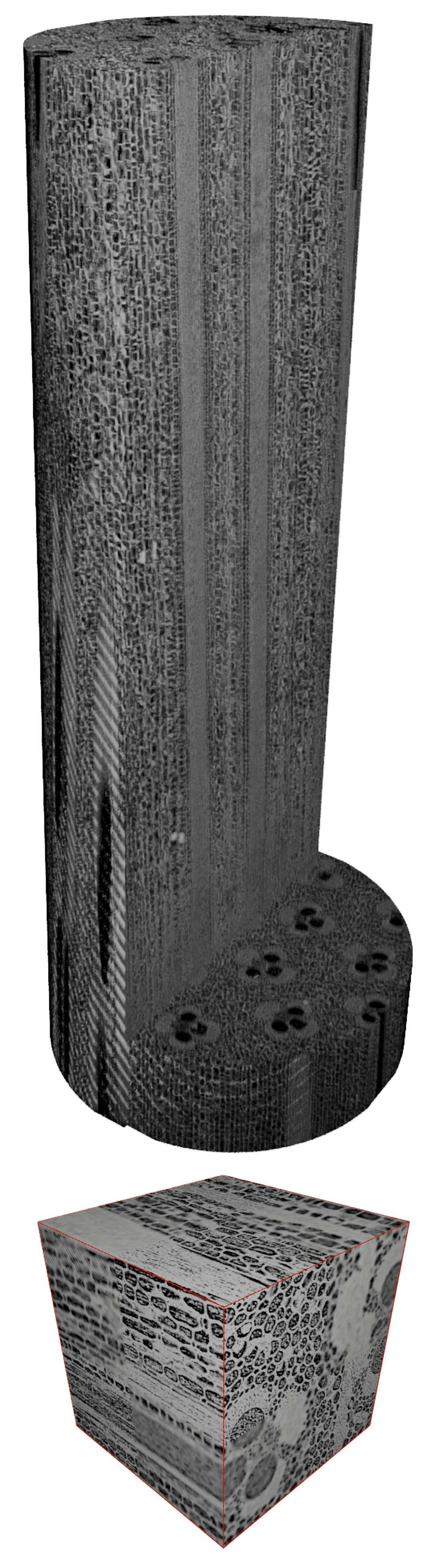}
        \label{fig:sub1}
    \end{subfigure}
    \hfill
    \begin{subfigure}[b]{0.09\textwidth}
        \centering
        \includegraphics[width=\textwidth]{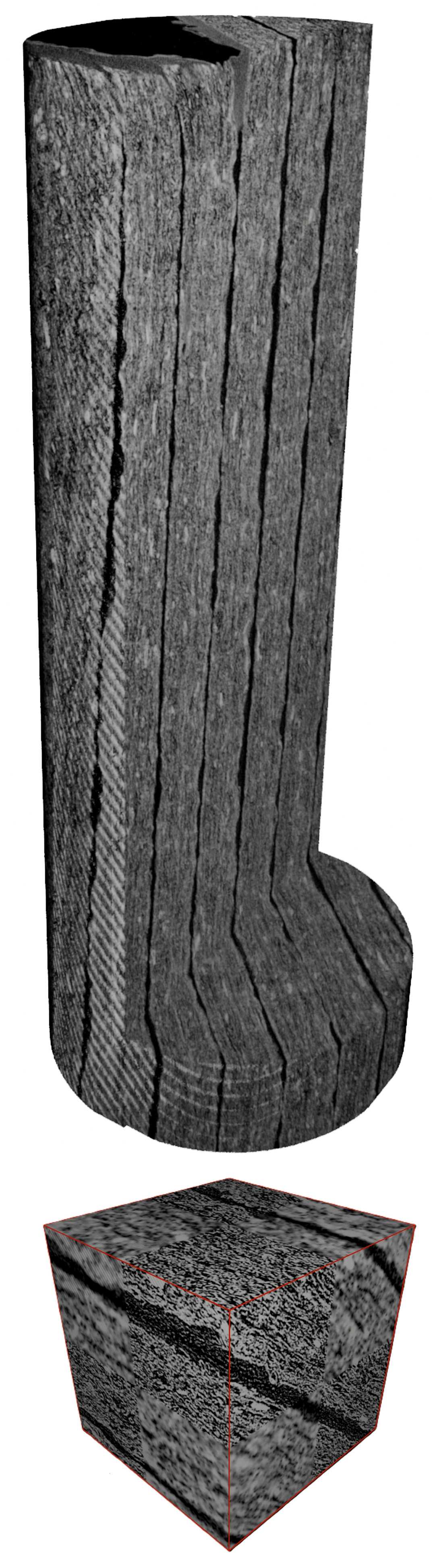}
        \label{fig:sub2}
    \end{subfigure}
    \hfill
    \begin{subfigure}[b]{0.09\textwidth}
        \centering
        \includegraphics[width=\textwidth]{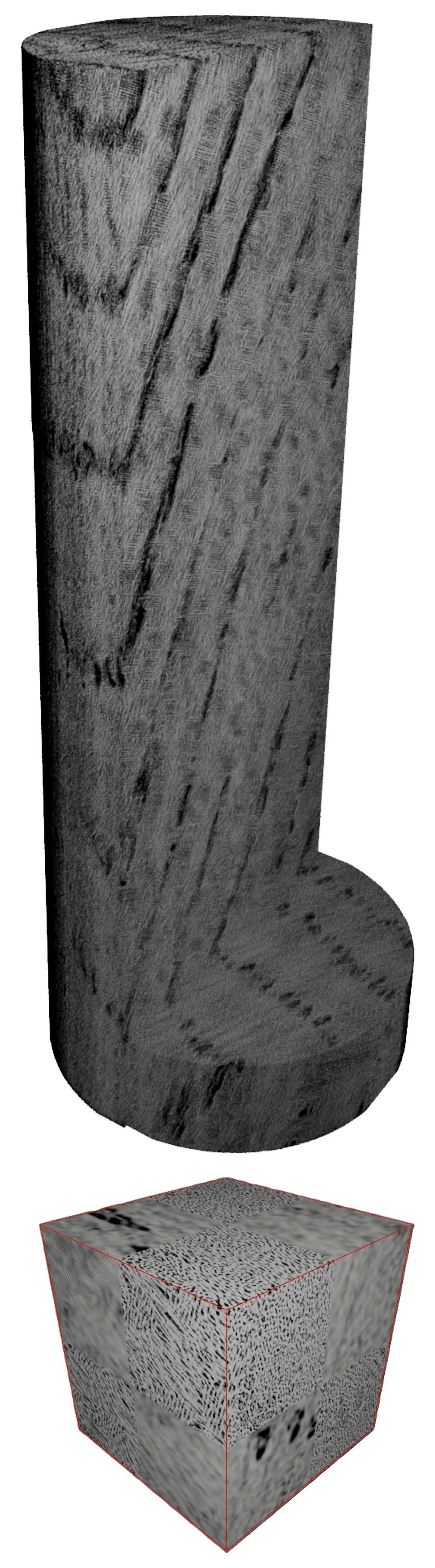}
        \label{fig:sub2}
    \end{subfigure}
    \hfill
    \begin{subfigure}[b]{0.09\textwidth}
        \centering
        \includegraphics[width=\textwidth]{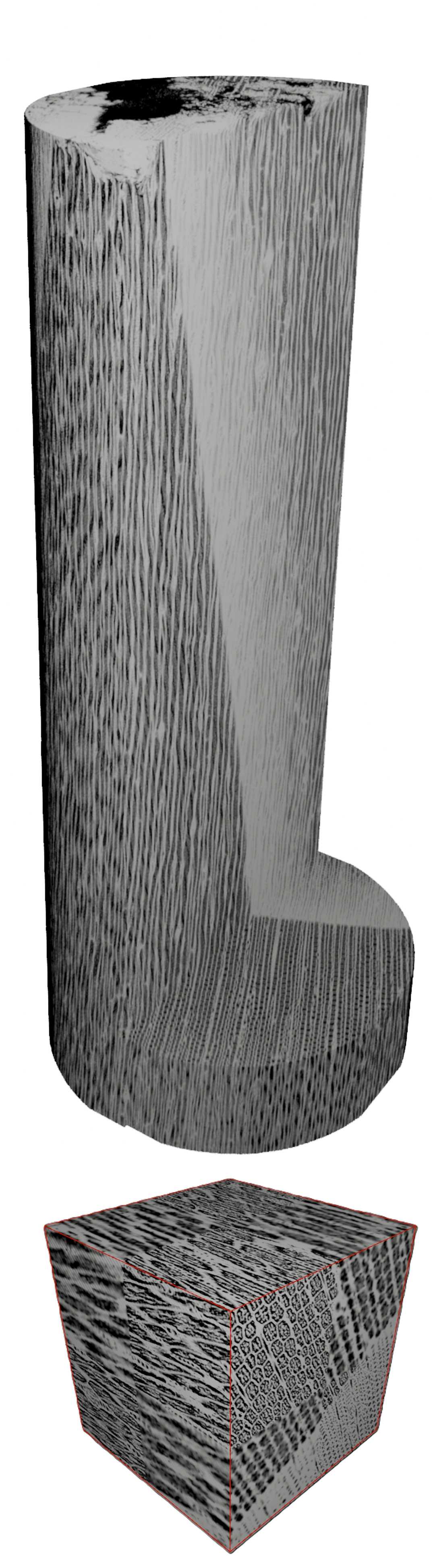}
        \label{fig:sub2}
    \end{subfigure}
    \hfill
    \begin{subfigure}[b]{0.09\textwidth}
        \centering
        \includegraphics[width=\textwidth]{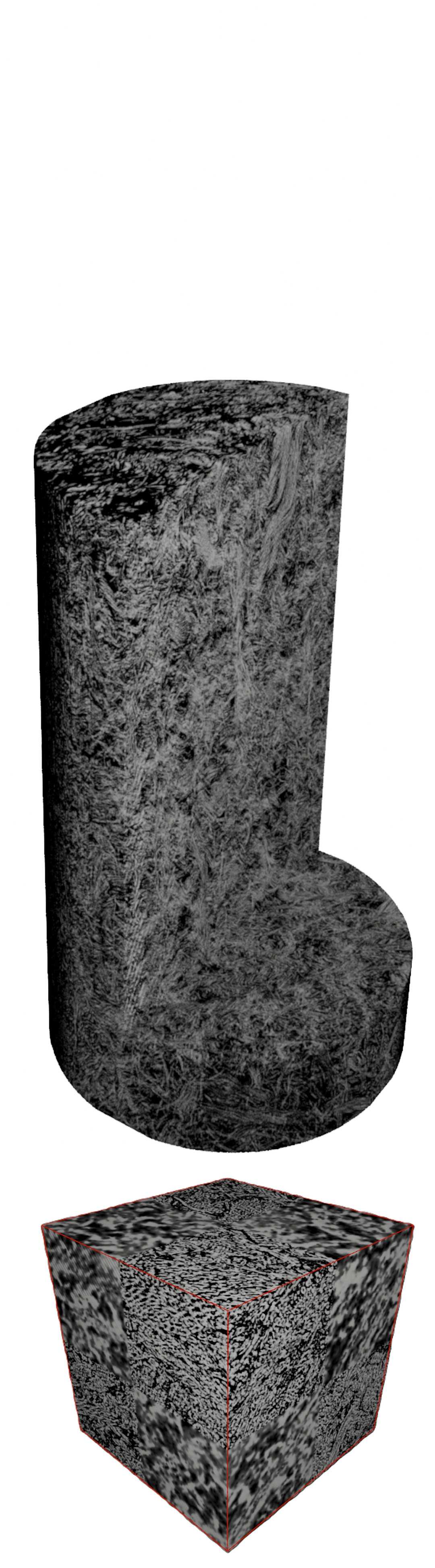}
        \label{fig:sub2}
    \end{subfigure}
    \hfill
    \begin{subfigure}[b]{0.09\textwidth}
        \centering
        \includegraphics[width=\textwidth]{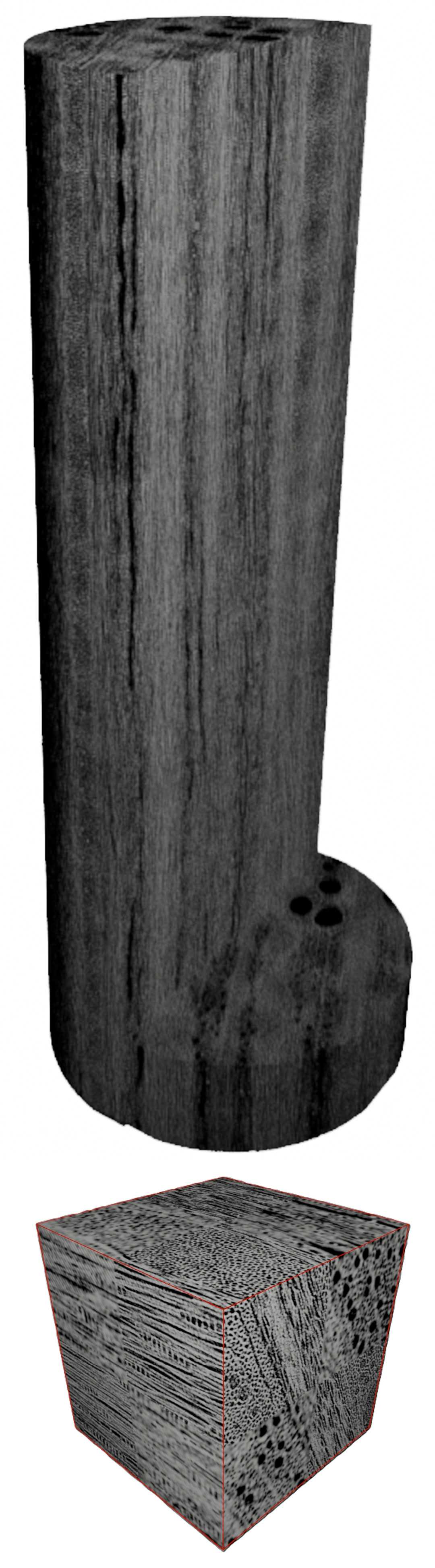}
        \label{fig:sub2}
    \end{subfigure}
    \hfill
    \begin{subfigure}[b]{0.09\textwidth}
        \centering
        \includegraphics[width=\textwidth]{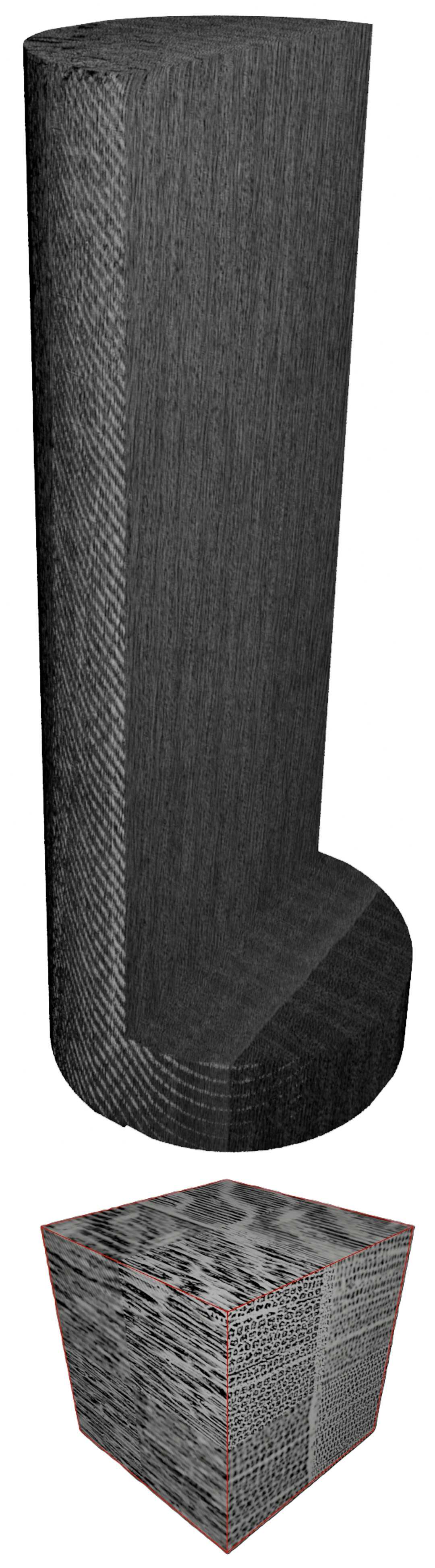}
        \label{fig:sub2}
    \end{subfigure}
    \hfill
    \begin{subfigure}[b]{0.09\textwidth}
        \centering
        \includegraphics[width=\textwidth]{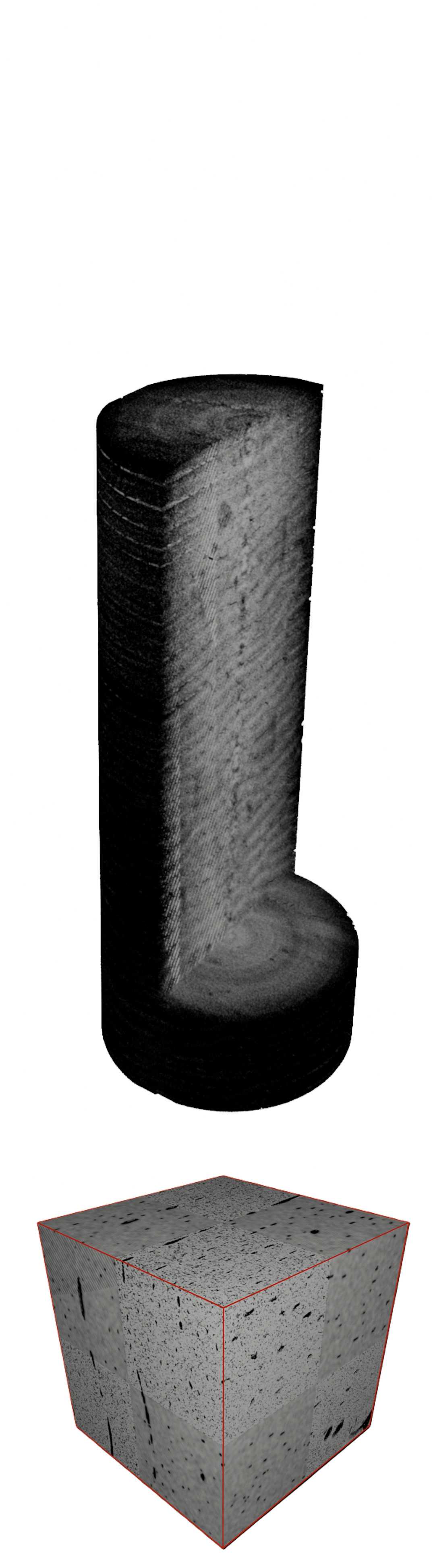}
        \label{fig:sub2}
    \end{subfigure}
    \hfill
    \begin{subfigure}[b]{0.09\textwidth}
        \centering
        \includegraphics[width=\textwidth]{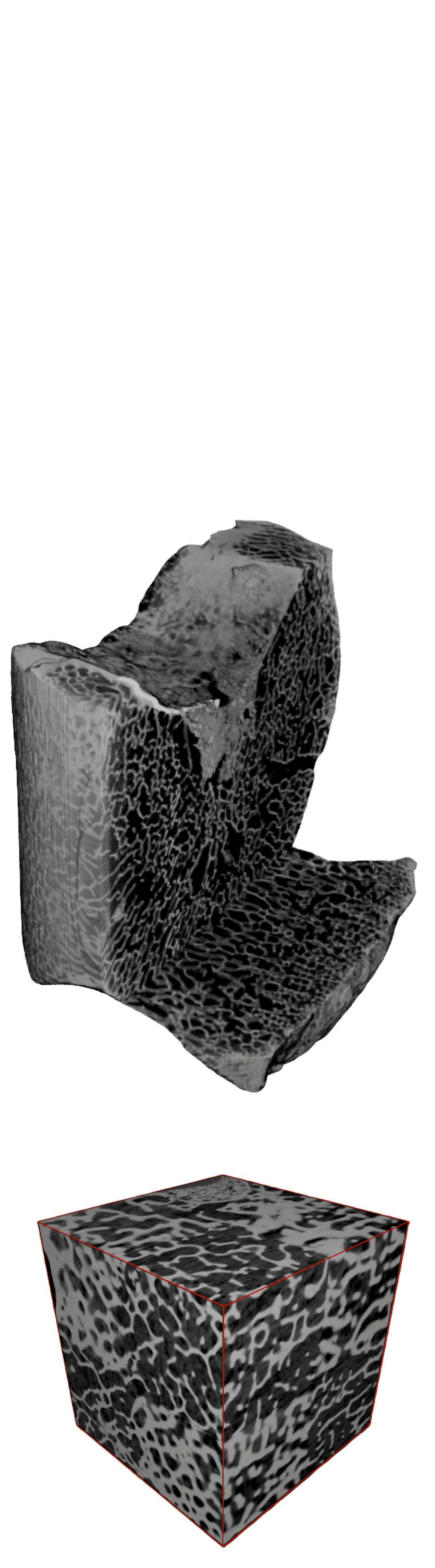}
        \label{fig:sub2}
    \end{subfigure}
    \hfill
    \begin{subfigure}[b]{0.09\textwidth}
        \centering
        \includegraphics[width=\textwidth]{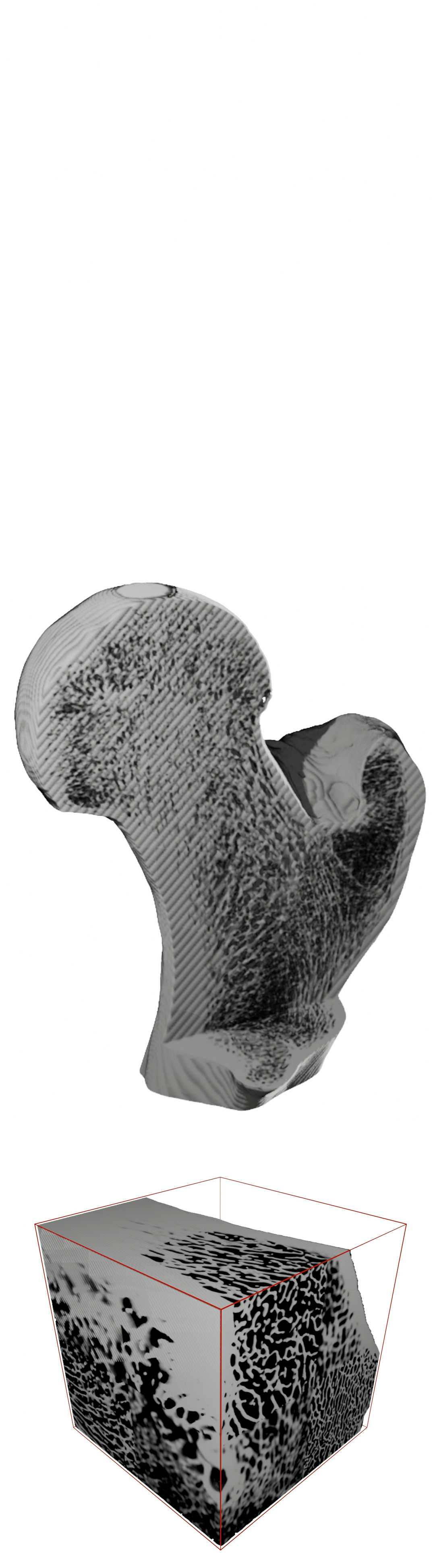}
        \label{fig:sub2}
    \end{subfigure}

    \caption{\datasetname sample overview, including zoom-ins shown in red-framed cubes. From left to right: bamboo, cardboard, elm, larch, MDF, oak, cypress, animal bone, human vertebrae, and human femur. High- and low-resolution scans are interleaved in the zoom-in cubes.}
    \label{fig:combined}
\end{figure*}

When trained on real low-resolution scans, i.e., volumes recorded at different resolutions, the predicted structures change dramatically, losing the high-frequency details observed in reconstructions obtained using downsampled input data (\cref{fig:intro_overview_vodasure}). This raises a question: Are current SR methods capable of recovering structures that vanish at low resolution, or do they simply predict plausible averages? 

To address this question, we introduce \textbf{VoDaSuRe} -- the \textbf{Vo}lumetric \textbf{Da}taset for \textbf{Su}per-\textbf{Re}solution, a large-scale CT dataset designed for SR research. \datasetname contains samples scanned at multiple resolutions using the same micro-CT scanning setup, along with downsampled volumes for benchmarking. Using \datasetname, we investigate the performance discrepancy of SR models trained on downsampled data vs. registered data scanned in actual low-resolution. 

Our results reveal the presence of a domain shift in SR predictions: models trained on downsampled data recover fine structures accurately, while models trained on scanned low-resolution data produce unrealistic, spatially averaged predictions. Applying models trained on downsampled data to real low-resolution scans surprisingly produces volumes that appear structurally plausible, yet lack precision compared to predictions obtained from training using downsampled data. These findings highlight the limitations of current SR approaches and the need for datasets with physically acquired LR data that enable scientifically relevant progress.

\datasetname spans diverse structural complexity, including wood, composite material, and bone (human and animal). Bone volumes exhibit high contrast and smooth variations, whereas wood-based samples range from regular solid wood to chaotic fiber composites with extremely fine canal structures. In addition to high structural complexity, \datasetname is the largest volumetric SR dataset in terms of total voxel count with paired multi-resolution data, with $16$ paired scans ($32$ in total), comprising $\sim194.0 \times 10^9$ voxels.

\datasetname is intended as a research benchmark for studying volumetric SR under realistic acquisition conditions. Although we do not claim direct transferability to clinical MRI/CT, \datasetname includes bone scans relevant to medical analysis tasks such as estimating bone volume fraction and lacunar statistics. 
Clinical photon-counting detectors \cite{greffier2025photon,quintiens2024photon} and cone-beam CT \cite{ko2024role,siddall2024emerging,yu2022clinical} for imaging fine bone structures and soft tissue have also emerged, which motivates studying SR for complex microstructures.

\noindent
\textbf{Our contributions are:}
\begin{itemize}
    \item We introduce \datasetname, a large-scale volumetric SR dataset with paired high- and low-resolution scans and synthetically downsampled versions for comparison.
    \item We benchmark state-of-the-art CNN- and ViT-based SR methods across multiple volumetric datasets, including medical datasets and \datasetname, revealing large domain gaps between downsampled and real LR data settings.
    \item We implement a data pipeline based on the OME-Zarr format for efficient out-of-core sampling of 3D patches.
    \item We release \datasetname and our code publicly to enable reproducible SR research.
\end{itemize}

\begin{table*}[t]
\renewcommand{\arraystretch}{1.0}

\begin{subtable}[t]{\linewidth}
\centering
\begin{scriptsize}

\begin{tabular*}{\linewidth}{@{\extracolsep{\fill}}cccccccc@{}}
\toprule
Dataset & \#samples & \#gigavoxels & Volume shape & Domain & Modalities & Accessibility & Paired res. \\ \midrule
NAMIC \cite{NAMIC_wiki} & 20 & $\sim0.2$ & $256\times256\times176$ & Brain MRI & T1w, T2w, fMRI, DTI & On request & \xmark \\ 
Kirby 21 \cite{Landman_Kirby21_dataset} & 21 & $\sim0.3$ & $256\times256\times180$ & Brain MRI & T2w, FLAIR, DTI & Public & \xmark \\ 
IXI\footnote{\url{https://brain-development.org/ixi-dataset/}} & 600 & $\sim5.9$ & $256\times256\times(144\text{-}150)$ & Brain MRI & T1w, T2w, PD & Public & \xmark \\ 
fastMRI \cite{zbontar2018fastmri} & 8400 & $\sim11.0$ & $256\times256\times20$ (mean) & Brain \& knee MRI & T1w, T2w, FLAIR & Public & \xmark \\ 
BraTS 2023 \cite{Menze_BraTS_GLI} & 1470 & $\sim13.1$ & $240\times240\times155$ & Brain MRI (Glioma) & T1w, T2w & On request & \xmark \\ 
LiTS \cite{bilic_LITS} & 130 & $\sim17.4$ & $512\times512\times(74\textbf{-}987)$ & Liver CT & Clinical CT & Public & \xmark \\ 
HCP 1200 \cite{Van_HCP_1200_dataset} & 1113 & $\sim29.2$ & $320\times320\times256$ & Brain MRI & T1w, T2w & Public & \xmark \\ 
LIDC-IDRI \cite{Armato_LIDC_IDRI} & 1010 & $\sim104.3$ & $512\times512\times(65\textbf{-}764)$ & Thorax CT & Clinical CT & Public & \xmark \\ 
CTSpine1K \cite{deng_CTSpine1K} & 1005 & $\sim134.9$ & $512\times512\times(85\textbf{-}736)$ & Spine CT & Clinical CT & Public & \xmark \\ 

\midrule
\citet{solidbattery2023} & 1 (2) & $\sim0.1$ & $ 512 \times 512 \times 200$ & Materials & Lab- \& synchrotron CT & On request &  \cmark \\ 
\citet{li2022deep} & 61 (122) & $\sim0.9$ & $512 \times 512 \times 28$ & Brain MRI & T1w, T2w, FLAIR & On request &  
\cmark \\
\citet{chu2025paired} & 20 (40) & $\sim2.2$ & $480 \times 512 \times 224$ & Brain MRI & T1w, T2w & Unavailable &  
\cmark \\
WAND \cite{mcnabb2025wand} & 170 (846) & $\leq19.4$ & $\leq 320\times320\times224$ & Brain MRI & T1w, T2w, fMRI, TRUST & Public &  \cmark \\ 
\citet{klos2025super} & 1 (6) & $\sim20.5$ & $1181\times 1695\times 1695$ (mean) & Materials & Lab-CT & Public &  
\cmark \\ 
\citet{synchrotronfiber2023} & 1 (2) & $\sim28.3$ & $2560 \times 2560 \times 2160$ (mean) & Materials & Synchrotron CT & On request &  
\cmark \\ 
RPLHR-CT \cite{RPLHR_CT} & 250 (500) & $\sim52.0$ & $512 \times 512\times(191\text{-}396)$ & Thorax CT & Clinical CT & On request &  \cmark \\ 
I13-2 XCT \cite{green2025three} & 1 (4) & $\sim53.2$ & $2510\times2510\times2110 $ & Materials & Synchrotron CT & Public &  \cmark \\ 
FACTS \cite{Bardenfleth_FACTS} & 13 (26) & $\sim57.6$ & $1014 \times 1372 \times 1584$ (mean) & Femur CT & Lab- \& clinical CT & Public &  \cmark \\ 

\midrule
VoDaSuRe (ours) & 16 (32) & $\sim\mathbf{194.0}$ & $\mathbf{3330 \times 1820 \times 1870}$ (mean)\footnote{The volume shape of the samples in VoDaSuRe varies, however all HR scans are chunked in cubes of $160^3$ voxels.}  & Medical/Materials & Lab CT & Public &  \cmark \\ 
\bottomrule
\end{tabular*}

\end{scriptsize}
\label{tab:ablation_levels}
\end{subtable}

\caption{Overview of volumetric image datasets, including single-resolution and paired multi-resolution datasets. In terms of total corresponding LR-HR voxel pairs and mean volume shape, \datasetname is several times larger than existing paired resolution 3D datasets.}
\label{tab:dataset_table}
\end{table*}
\section{Related work}
\label{sec:related_work}

\textbf{Super-resolution for volumetric images}.
A family of SR approaches for 3D volumes adopts slice-wise strategies \cite{Chen_HAT, Zhang_RCAN}, where each slice is upscaled independently to increase in-plane resolution. While computationally efficient, these approaches ignore cross-plane information and risk discontinuities across volumetric predictions. In contrast, fully volumetric methods use 3D convolutional \cite{Wang_ESRGAN, Wang_EDDSR} or ViT layers \cite{Forigua_SuperFormer, Hoeg_MTVNet} to learn spatial correlations in all dimensions, improving accuracy at higher computational cost.

As reseach interest in volumetric SR expands, various approaches have emerged. Axial SR methods increase resolution in the slice direction of LR MRI while preserving in-plane resolution \cite{Ge_ResVoxGAN, Ji_3DSR_survey, RPLHR_CT, Wang_ASFT}. Arbitrary scale SR methods produce continuously upscaled volumetric images using implicit neural representations \cite{Wu_ArSSR, Zhu_MIASSR, Li_McASSR, mcginnis2023single}, whereas multi-contrast SR methods exploit images from multiple imaging modalities to enrich feature extraction \cite{Ji_3DSR_survey, Li_McASSR, mcginnis2023single}. 

A common practice in most SR research is the use of synthetically generated data, where LR volumes are produced by downsampling their HR counterparts. This practice makes the SR problem easier than when using LR data physically acquired at reduced resolution. True LR acquisition differs fundamentally from downsampling: it often provides higher contrast and better signal-to-noise ratios, but may introduce acquisition or reconstruction artifacts. To be practical, SR methods must reconstruct HR images despite these artifacts. Therefore, data for developing SR methods must have the same properties as data scanned in different resolutions. Despite the proposal of sophisticated degradation models, progress towards improving the generalization of volumetric SR in real-world scenarios remains limited by the lack of paired multi-resolution volumetric datasets.

\textbf{Volumetric datasets}.
Although volumetric datasets vary in acquisition methods, resolution scales, and application domains, medical imaging datasets remain the most common data category for volumetric SR \cite{Ji_3DSR_survey}. The use of SR for medical images is motivated by benefits such as higher diagnostic accuracy, better treatment planning, shorter acquisition times, and reduced radiation exposure \cite{Chen_mDCSRN_A, Chen_mDCSRN_B, Chen_mDCSRN_C, Ji_3DSR_survey}.

The most widely used medical datasets used for SR are listed in the top part of \cref{tab:dataset_table}. NAMIC \cite{NAMIC_wiki} was used for SR in \cite{iwamoto2022unsupervised, Pham_ReCNN}, Kirby 21 \cite{Pham_SRCNN3D} has been used for SR in \cite{Wang_EDDSR, Zhang_CFTN, Pham_ReCNN, Wang_FASR, Du_DCED, Li_MFER, Hoeg_MTVNet}, IXI (Information eXtraction from Images) \cite{li2021volumenet} has been used for SR in \cite{Zhang_SERAN, Wang_W2AMSN, Ji_3DSR_survey, Wang_InverseSR, Li_MFER, Hoeg_MTVNet}, fastMRI \cite{zbontar2018fastmri} was used for SR in \cite{Li_McASSR}, BraTS (Brain Tumor Segmentation Challenge) \cite{Pham_SRCNN3D} was used for SR in \cite{Wang_EDDSR, Li_McASSR, Zhu_MIASSR, Hoeg_MTVNet}, and HPC (Human Connectome Project) \cite{Chen_mDCSRN_A, Chen_mDCSRN_B, Chen_mDCSRN_C} has been used for SR in \cite{Lu_novel, Forigua_SuperFormer, li2021volumenet, Wu_ArSSR, zhou2022blind, Wang_MRGD, Hoeg_MTVNet}. Additionally, ADNI (Alzheimer’s Disease Neuroimaging Initiative) was employed for SR in \cite{Sanchez_SRGAN3D, Lu_novel}.

Despite their popularity, most medical datasets have relatively low in-plane resolution of $\leq 320^2$ voxels. 
Recently, higher resolution medical datasets with an in-plane size of $512^2$ voxels have emerged, including CTSpine1K \cite{deng_CTSpine1K}, LiTS \cite{bilic_LITS}, KiTS \cite{heller2023kits21}, and LIDC-IDRI \cite{Armato_LIDC_IDRI}. However, these datasets have not been widely used in volumetric SR.


Acquiring 3D data at multiple resolutions is expensive, slow, and increases radiation exposure, so most 3D datasets are available only at a single resolution. Consequently, LR data for SR training must be generated synthetically. Typical downsampling methods involve Gaussian smoothing followed by interpolation (e.g., cubic or linear), used for slice-wise, volumetric, and axial SR. To better approximate MRI acquisition, k-space truncation has been proposed \cite{Chen_mDCSRN_A, Chen_mDCSRN_B, Chen_mDCSRN_C}, which removes high-frequency components to introduce aliasing while preserving spatial dimensions. Expanding on this, \citet{ayaz2024effective} proposed a more comprehensive degradation approach to simulate LR MRI acquisition. 

\begin{figure*}[t]
  \centering
  \includegraphics[width=0.98\linewidth]{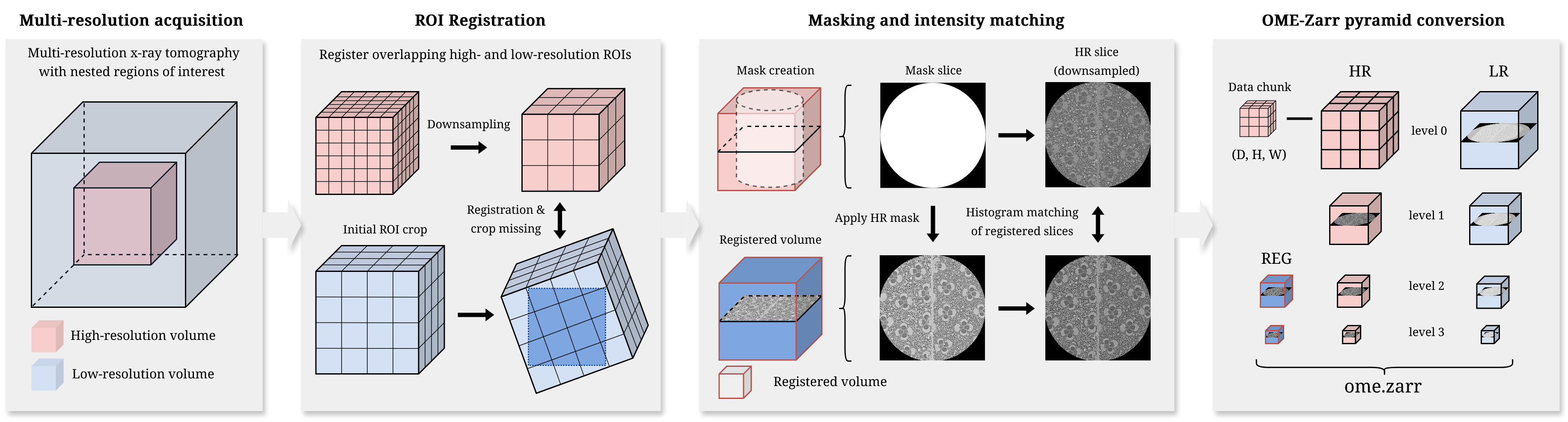}
    \caption{Illustration of our data curation pipeline for \datasetname. We collect multi-resolution nested CT scans of the same sample, after which we crop and register the LR data to the downsampled HR volumes. LR and HR volumes are masked and their intensity histograms are matched. All scans are saved to OME-Zarr with up to four resolution levels, using separate groups for HR, LR, and registered data.} 
  \label{fig:VoDaSuRe_preprocessing}
\end{figure*}

\textbf{Paired image datasets for volumetric SR}.
Despite the proposal of more realistic SR degradation models for generating LR data from HR scans, such methods fail to capture CT-related artifacts such as beam hardening, motion, and ring artifacts. Instead, the actual differences are best obtained by scanning paired LR-HR images.


The datasets listed in the lower part of \cref{tab:dataset_table} include scans acquired at multiple resolutions. These comprise:  
lab-CT and synchrotron CT of batteries by \citet{solidbattery2023};  
paired brain MRI scans from different scanners by \citet{li2022deep};  
multi-modal brain MRI for cross-modal SR by \citet{chu2025paired};  
the WAND dataset of brain MRI with varying modalities and protocols \cite{mcnabb2025wand};  
a recent lab-CT dataset of a diamond-like lattice cuboid scanned at six resolutions by \citet{klos2025super};  
the RPLHR-CT dataset with 250 paired chest CT volumes \cite{RPLHR_CT};  
the I13-2 XCT synchrotron CT dataset of a Zinc-Doped Zeolite sample at four resolutions \cite{green2025three};  
and the Femur Archaeological CT Super-resolution (FACTS) dataset with 12 femurs scanned using clinical and lab-CT \cite{Bardenfleth_FACTS}. In addition to these, \citet{synchrotronfiber2023} curated a paired synchrotron CT dataset of fiber composites.

Although these works enable more realistic SR tasks, the volumes within these datasets are either 1) small-scale ($\leq 512^3$), 2) span a narrow set of microstructural domains, 3) provide a small number of volumes, 4) or are not easy to download. Their limited size and narrow domain coverage prevent fair and reproducible method comparison. In contrast, VoDaSuRe provides registered, real multi-resolution volumetric scans acquired across multiple structural domains, offering the first large-scale benchmark for studying genuine resolution enhancement in volumetric imaging.

\section{\datasetname dataset}
\label{sec:data}

The \datasetname dataset consists of multi-resolution X-ray CT volumes of 16 biological and non-biological samples spanning diverse volumetric microstructures. In terms of total voxel for which paired LR data have been acquired, \datasetname is the largest volumetric dataset to feature both synthetically downsampled \textit{and} physically acquired, co-registered LR scans obtained using the same scanner setup. Unlike multi-modality datasets, where discrepencies between low and high resolution arise from different imaging setups (e.g., micro-CT vs.\ clinical CT), \datasetname isolates the effects of pure resolution difference, allowing the study of realistic degradation processes and enabling training of models that generalize across real-world resolution scales. In terms of samples, our dataset includes four human femurs and four vertebrae, animal bone (ox), wood samples from five tree species, and composites including medium-density fiberboard (MDF) and cardboard laminate (\cref{fig:combined}). An overview of all samples, including scanner type, resolution levels, and volume dimensions for HR, LR, and co-registered scans, is provided in the supplementary material.

\begin{figure*}[t]
  \centering
  \begin{subfigure}{0.96\linewidth}
    \includegraphics[width=1.0\linewidth]{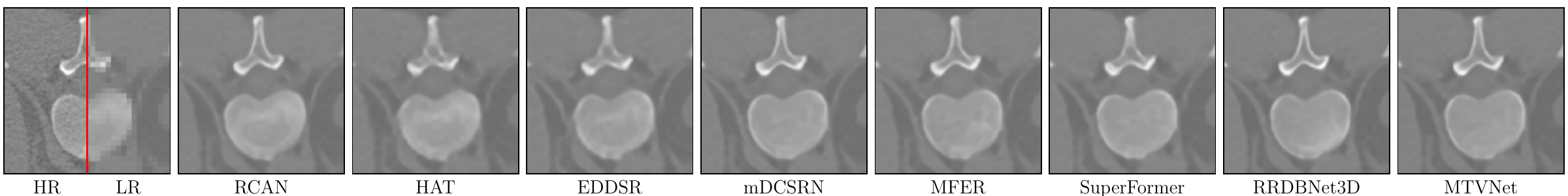}
  \end{subfigure}
  \hfill
  \begin{subfigure}{0.96\linewidth}
    \includegraphics[width=1.0\linewidth]{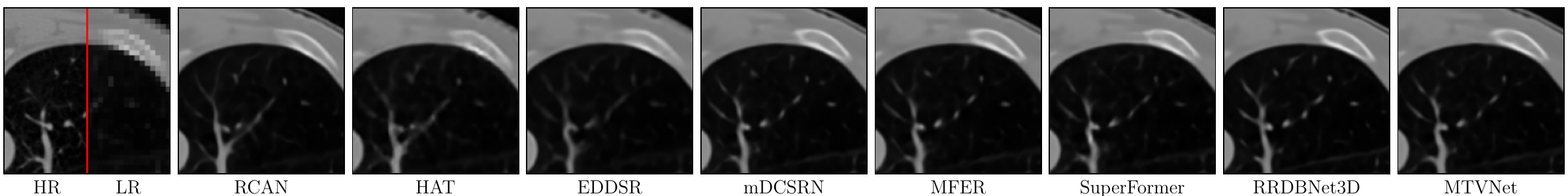}
  \end{subfigure}
  \hfill
  \begin{subfigure}{0.96\linewidth}
    \includegraphics[width=1.0\linewidth]{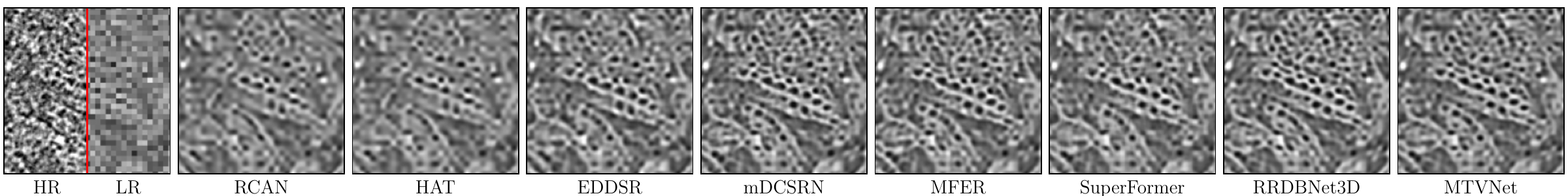}
  \end{subfigure}
  \hfill
  \begin{subfigure}{0.96\linewidth}
        \includegraphics[width=1.0\linewidth]{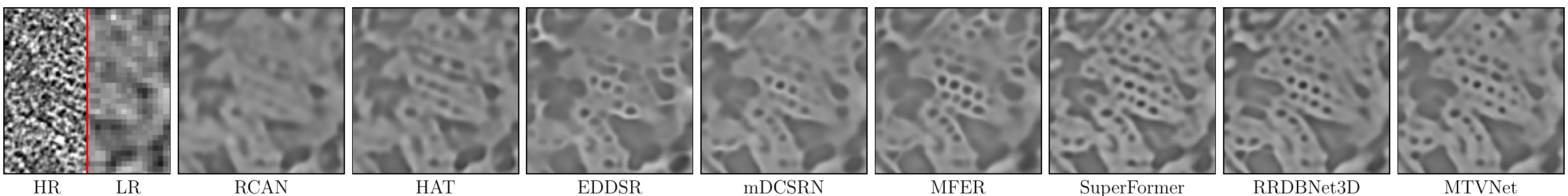}
  \end{subfigure}
  \hfill
  \begin{subfigure}{0.96\linewidth}
    \includegraphics[width=1.0\linewidth]{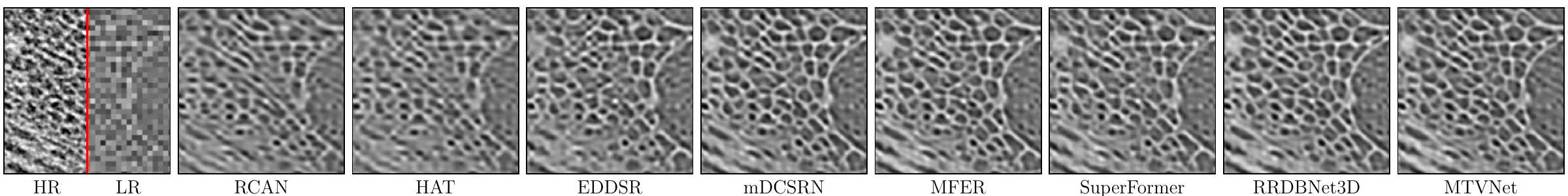}
  \end{subfigure}
  \hfill
  \begin{subfigure}{0.96\linewidth}
    \includegraphics[width=1.0\linewidth]{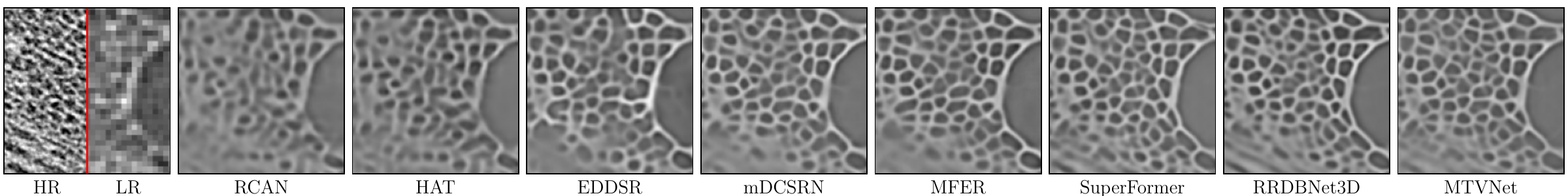}
  \end{subfigure}
  
  \caption{Visual comparison of SR predictions on CTSpine1K, LIDC-IDRI and \datasetname at scale $4\times$. From top to bottom: CTSpine1K, LIDC-IDRI, \datasetname (downsampled), and \datasetname (registered) -- two examples. The LR inputs and corresponding HR ground truth images are shown on the left, separated by the red line.}
  \label{fig:model_visualizations}
\end{figure*}

\begin{table*}[t]
\centering
\renewcommand{\arraystretch}{1.00}

\resizebox{\textwidth}{!}{%
\begin{footnotesize}
\begin{NiceTabular}{cccccc}[]
\toprule
\Block{2-1}{\normalsize Method}
& \Block{2-1}{\normalsize CTSpine1K} 
& \Block{2-1}{\normalsize LiTS} 
& \Block{2-1}{\normalsize LIDC-IDRI} 
& \Block{2-1}{\normalsize VoDaSuRe \\\normalsize (Downsampled)} 
& \Block{2-1}{\normalsize VoDaSuRe \\\normalsize (Registered)} \\
\\
\midrule
\normalsize Scale $2\times$ 
& \footnotesize{PSNR / SSIM / NRMSE / LPIPS} 
& \footnotesize{PSNR / SSIM / NRMSE / LPIPS} 
& \footnotesize{PSNR / SSIM / NRMSE / LPIPS} 
& \footnotesize{PSNR / SSIM / NRMSE / LPIPS}
& \footnotesize{PSNR / SSIM / NRMSE / LPIPS} \\ \midrule

HAT & 35.94 / .9258 / .0430 / .0571 & 38.11 / .9713 / .0339 / .0257 & 33.92 / .9062 / .0419 / .0614 & 23.32 / .8921 / .1638 / .1571 &  17.44 / .4755 / .3377 / .4661 \\ 
RCAN & 36.62 / .9282 / .0402 / .0505 & 39.04 / .9736 / .0309 / .0218 & 35.01 / .9111 / .0379 / .0549 & 23.28 / .8687 / .1654 / .1575 & 17.40 / .4691 / .3337 / .4807 \\ 
EDDSR & 37.04 / .9292 / .0380 / .0597 & 39.87 / .9752 / .0274 / .0268 & 34.40 / .9096 / .0395 / .0651 & 24.86 / .9066 / .1445 / .1563 & 17.63 / .4755 / .3277 / .4527 \\ 
SuperFormer & 38.24 / .9330 / .0338 / .0495 & 41.45 / .9780 / .0229 / .0197 & 37.03 / .9162 / .0311 / .0526 & 25.00 / .9090 / .1429 / .1398 & 18.17 / .5246 / .3089 / .3958 \\ 
MFER & 39.05 / .9325 / .0317 / .0458 & 43.14 / .9776 / .0194 / .0168 & 38.82 / .9160 / .0274 / .0462 & 25.30 / .9093 / .1404 / .1457 &  17.98 / .5152 / .3184 / .4176 \\ 
mDCSRN & 38.89 / .9352 / .0317 / .0430 & 42.58 / .9798 / .0203 / .0172 & 37.84 / .9190 / .0289 / .0463 & 25.17 / .9089 / .1438 / .1457 & 17.97 / .5110 / .3149 / .4220 \\ 
MTVNet & 39.27 / .9355 / .0307 / .0436 & 43.16 / .9801 / .0192 / .0164 & 39.36 / .9215 / .0261 / .0430 & 24.82 / .9066 / .1479 / .1434 & 18.21 / .5249 / .3080 / .3898 \\  

RRDBNet3D & $\textbf{39.88}$ / $\textbf{.9387}$ / $\textbf{.0288}$ / $\textbf{.0400}$ & $\textbf{44.81}$ / $\textbf{.9821}$ / $\textbf{.0161}$ / $\textbf{.0141}$ & $\textbf{40.47}$ / $\textbf{.9256}$ / $\textbf{.0239}$ / $\textbf{.0382}$ & $\textbf{25.50}$ / $\textbf{.9106}$ / $\textbf{.1397}$ / $\textbf{.1378}$ &  $\textbf{18.25}$ /  $\textbf{.5377}$ /  $\textbf{.3057}$ / $\textbf{.3864}$ \\ 
\bottomrule
\end{NiceTabular}
\end{footnotesize}
}

\vspace{0.1cm}

\resizebox{\textwidth}{!}{%
\begin{footnotesize}
\begin{NiceTabular}{cccccc}[]
\toprule
\Block{2-1}{\normalsize Method}
& \Block{2-1}{\normalsize CTSpine1K} 
& \Block{2-1}{\normalsize LiTS} 
& \Block{2-1}{\normalsize LIDC-IDRI} 
& \Block{2-1}{\normalsize VoDaSuRe \\\normalsize (Downsampled)} 
& \Block{2-1}{\normalsize VoDaSuRe \\\normalsize (Registered)} \\
\\\midrule
\normalsize Scale $4\times$ 
& \footnotesize{PSNR / SSIM / NRMSE / LPIPS} 
& \footnotesize{PSNR / SSIM / NRMSE / LPIPS} 
& \footnotesize{PSNR / SSIM / NRMSE / LPIPS} 
& \footnotesize{PSNR / SSIM / NRMSE / LPIPS}
& \footnotesize{PSNR / SSIM / NRMSE / LPIPS} \\ \midrule

HAT & 30.44 / .8391 / .0839 / .1776 & 31.72 / .9106 / .0732 / .1083 & 29.50 / .8270 / .0721 / .1717 &  16.61 / .5032 / .3553 / .3855 & 15.41 / .3543 / .4526 / .4986 \\ 
RCAN & 31.38 / .8512 / .0757 / .1635 & 32.83 / .9229 / .0652 / .0854 & 30.71 / .8417 / .0634 / .1550 & 16.95 / .5512 / .3434 / .3722 & 15.37 / .3235 / .4146 / .5076 \\ 
EDDSR & 31.57 / .8530 / .0729 / .1738 & 33.20 / .9255 / .0610 / .0978 & 30.12 / .8433 / .0675 / .1679 & 18.15 / .6222 / .3051 / .3488 & 15.82 / .3995 / .3889 / .4650 \\ 
SuperFormer & 33.95 / .8674 / .0561 / .1539 & 35.58 / .9396 / .0457 / .0747 & 33.23 / .8616 / .0476 / .1378 &  18.53 / .6415 / .2931 / .3150 & \textbf{16.24} / \textbf{.4395} / \textbf{.3729} / $\textbf{.4250}$ \\ 
MFER & 34.36 / .8720 / .0540 / .1478 & 36.17 / .9439 / .0447 / .0687 & 33.40 / .8658 / .0526 / .1307 &  18.58 / .6518 / .2898 / .3283 & 15.99 / .4187 / .3851 / .4402 \\ 
mDCSRN & 34.77 / .8741 / .0512 / .1484 & 36.57 / .9461 / .0407 / .0684 & 33.92 / .8689 / .0437 / .1310 &  18.65 / .6666 / .2871 / $\textbf{.3121}$ &  16.11 / .4257 / .3781 / .4434 \\ 
MTVNet & 34.39 / .8742 / .0539 / .1464 & 36.25 / .9464 / .0428 / .0675 & 33.76 / .8696 / .0452 / .1261 &  18.81 / .6619 / .2888 / .3237 & 16.18 / .4204 / .3775 / .4408 \\ 
RRDBNet3D & $\textbf{35.57}$ / $\textbf{.8803}$ / $\textbf{.0472}$ / $\textbf{.1398}$ & $\textbf{37.86}$ / $\textbf{.9526}$ / $\textbf{.0353}$ / $\textbf{.0607}$ & $\textbf{35.26}$ / $\textbf{.8761}$ / $\textbf{.0387}$ / $\textbf{.1201}$ &  $\textbf{19.08}$ / $\textbf{.6779}$ / $\textbf{.2746}$ / .3250 & 16.22 / .4390 / .3729 / .4337 \\
\bottomrule
\end{NiceTabular}
\end{footnotesize}
}


\caption{Quantitative comparison of state-of-the-art SR models on datasets CTSpine1K, LiTS, LIDC-IDRI and \datasetname using downsampled and registered LR input data. The best performance metrics PSNR $\uparrow$ / SSIM $\uparrow$ / NRMSE $\downarrow$ / LPIPS $\downarrow$ are highlighted in \textbf{bold}.}
\vspace{-0.05in}
\label{tab:table_quantitative_results_in_domain}
\end{table*}

\begin{table}[t]
\centering
\renewcommand{\arraystretch}{1.00}

\resizebox{\linewidth}{!}{%
\begin{footnotesize}
\begin{NiceTabular}{ccc}[]
\toprule
\Block{1-1}{\normalsize Training} & \Block{1-2}{\normalsize{VoDaSuRe (Downsampled)}} & \\\midrule
\Block{1-1}{\normalsize Evaluation} & \Block{1-1}{\normalsize VoDaSuRe \\\normalsize(Registered) $2\times$} & \Block{1-1}{\normalsize VoDaSuRe \\\normalsize(Registered) $4\times$} \\\midrule
\normalsize Method & \footnotesize{PSNR / SSIM / NRMSE / LPIPS} & \footnotesize{PSNR / SSIM / NRMSE / LPIPS} \\ \midrule
HAT & 16.38 / .4806 / .3606 / .5220 & 14.58 / .3118 / .4445 / .5132 \\
RCAN & 16.38 / .4337 / .3644 / .5208 & 14.57 / .3429 / .4433 / .4894 \\
EDDSR & 16.62 / \textbf{.4847} / .3560 / .5101 & \textbf{15.18} / .3945 / .4202 / .4560 \\
SuperFormer & 16.54 / .4891 / .3579 / $\textbf{.5045}$ & 15.12 / .3895 / \textbf{.4191} / $\textbf{.4409}$ \\
MFER & 16.51 / .4812 / .3598 / .5063 & 14.89 / .3837 / .4296 / .4593 \\
mDCSRN & \textbf{16.77} / .4794 / .3576 / .5132 & 14.94 / \textbf{.4000} / .4266 / .4488 \\
MTVNet & 16.73 / .4689 / .3647 / .5148 & 14.72 / .3540 / .4407 / .4610 \\
RRDBNet3D &  16.74 / .4781 / \textbf{.3548} / .5092 & 14.94 / .3923 / .4221 / .4542 \\
\bottomrule
\end{NiceTabular}
\end{footnotesize}
}

\caption{Quantitative results for cross-domain experiments on \datasetname registered and downsampled at $2\times$ and $4\times$ upscaling.}
\vspace{-0.05in}
\label{tab:cross_domain_experiments}
\end{table}

\subsection{Data acquisition}

We faced an important choice when acquiring \datasetname: how large a resolution gap should separate the LR and HR scans. A gap that is too small leads to trivial SR tasks, whereas increasing the resolution beyond the smallest feature size only enlarges the data without adding meaningful structural information. Based on this trade-off, we selected a fixed resolution difference of $4\times$ between all HR and LR acquisitions. We then tuned the specific voxel-size ranges for each sample so that fine-scale features are fully resolved only in the HR scans, while coarser structures remain visible in both. This ensures that every LR–HR pair represents a non-trivial and informative SR task.


Because our samples contain microstructures at different spatial scales, we used two lab-CT scanners, each tuned for a specific resolution range.
Human vertebra and femur bone samples were scanned using a Nikon XT H 225 lab-CT scanner. The remaining samples were acquired using a Zeiss Xradia Versa 520 lab-CT scanner. In both setups, higher resolution scans were obtained by increasing sample-detector distance, which enlarges the projected cone angle to increase magnification at the cost of lower contrast. 

\subsection{Data curation}

Our data pipeline includes scanning, registration, masking, intensity matching, and OME-Zarr conversion, see \cref{fig:VoDaSuRe_preprocessing}.

\textbf{Initial processing}. All scans are percentile clipped to remove outliers. Volumes are then normalized to $[0; 65535]$ and cast to unsigned short. To enable SR tasks at multiple scales ($2\times$, $4\times$, $8\times$), we create pyramid volumes of both HR and LR scans using local mean downsampling, reducing resolution by a factor of $2$ up to a maximum of $8\times$. Foreground masks are created using either thresholding or UNet-based segmentation, depending on scan complexity.

\textbf{Registration}. Registration of LR/HR volumes is performed using ITK-Elastix \cite{itkelastix2023}. We initialize registration by pairing downsampled HR and LR scans with approximate voxel sizes. Each LR scan is coarsely cropped to the HR scan field of view, after which translational registration is done to achieve initial alignment. The alignment is then refined using affine registration, allowing small deformations of the LR volume to achieve voxel-level correspondence. Finally, the registered LR volumes are cropped to match the HR field of view, and voxels outside this region are masked.  

\textbf{Intensity matching}. To account for contrast differences between LR and HR scans, we match the intensities of each slice in the registered LR volumes. Specifically, we match the cumulative distribution function of all masked slices of each registered LR volume to their corresponding downsampled HR slices. This preserves the structure of each registered LR slice, while adjusting the relative intensities to match those in HR. This step is necessary in order to stabilize SR training, as the $L_{1}$-loss function used for optimization is highly sensitive to relative contrast differences. 


\textbf{OME-Zarr conversion and data loading}. The volumes in \datasetname are exceptionally large, requiring efficient storage and access. To mitigate memory constraints, all data were converted to the OME-Zarr format \cite{moore2023ome}, which extends Zarr with multi-resolution pyramids and OME-NGFF metadata, offering an ideal fit for SR tasks operating across scales. For each sample, we create an OME-Zarr image pyramids of the HR, LR, and registered volumes. Chunk sizes are empirically optimized for high I/O throughput and minimal cache misses when sampling HR/LR 3D patches. We further implement a PyTorch-compatible data loader that supports concurrent 3D patch sampling and augmentation, enabling out-of-core training on volumes exceeding system memory. Leveraging OME-Zarr’s hierarchical structure allows efficient access to arbitrary 3D patch sizes without pre-splitting or manual volume bookkeeping.

\begin{figure*}[t]
  \centering
  \begin{subfigure}{1.00\linewidth}
    \includegraphics[width=1.0\linewidth]{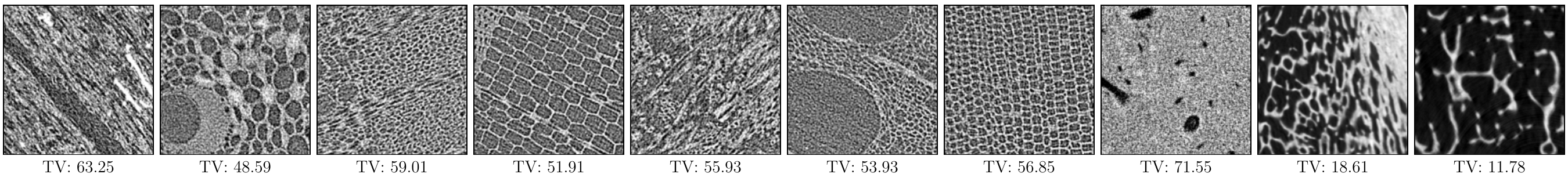}
  \end{subfigure}
  \hfill
  \begin{subfigure}{1.00\linewidth}
    \includegraphics[width=1.0\linewidth]{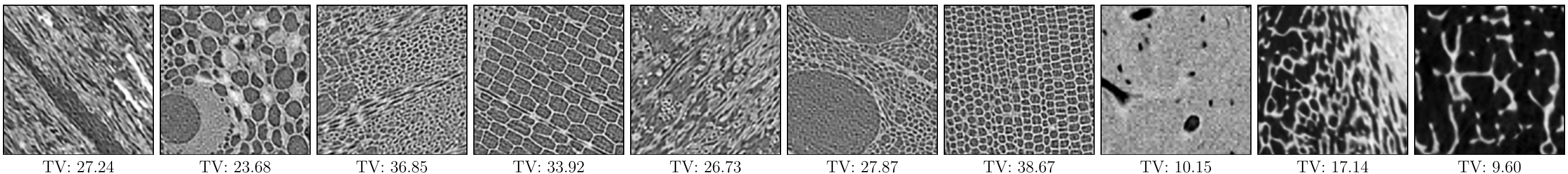}
  \end{subfigure}
   \hfill
  \begin{subfigure}{1.00\linewidth}
    \includegraphics[width=1.0\linewidth]{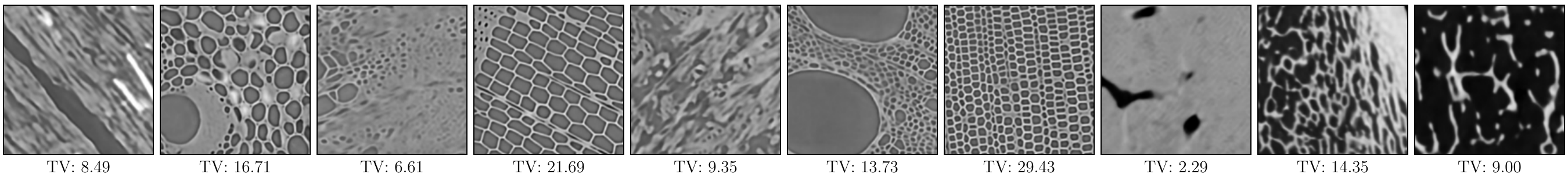}
  \end{subfigure}
  
  \caption{Visualizations from \datasetname. Top row: HR data, middle row: SR predictions using downsampled LR data, bottom row: SR predictions using real LR data. All outputs are obtained at $4\times$ upscaling using RRDBNet3D. Total variation (TV) is shown for each slice.}
  \label{fig:TV_visualizations}
\end{figure*}

\begin{figure}[t]
  \centering
  \includegraphics[width=1.0\linewidth]{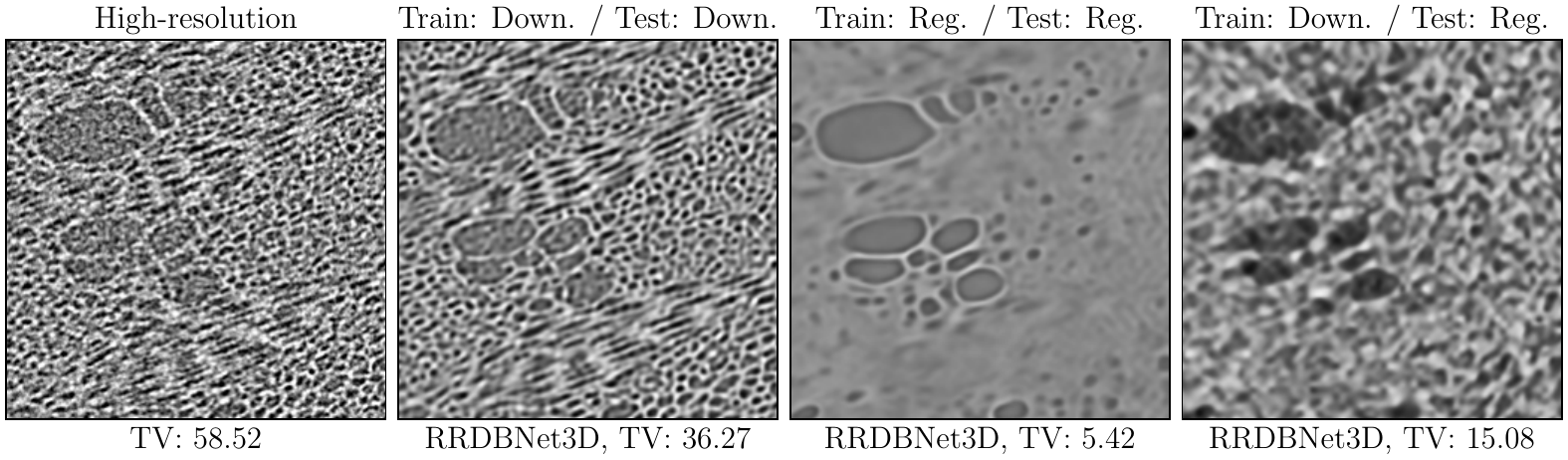}
    \caption{SR prediction obtained using different combinations of training/test data. From left to right: HR data, training/testing on downsampled data, training/testing on registered data, training on downsampled and testing using registered data. Predictions are obtained using RRDBNet3D at scale $4\times$.} 
  \label{fig:ablation_input}
\end{figure}

\section{Experiments}
\label{sec:experiments}

\textbf{Model selection}. We select eight recognized state-of-the-art SR methods, including six volumetric and two 2D approaches. The volumetric SR methods are: EDDSR \citep{Wang_EDDSR}, SuperFormer \citep{Forigua_SuperFormer}, MFER \citep{Li_MFER}, mDCSRN \citep{Chen_mDCSRN_C}, MTVNet \cite{Hoeg_MTVNet}, RRDBNet3D \cite{Wang_ESRGAN}, and the 2D SR methods are: RCAN \cite{Zhang_RCAN} and HAT \cite{Chen_HAT}. The version of MTVNet $L_{3}$ used in this study has fewer input features than in \cite{Hoeg_MTVNet}. 
We use the author's suggested parameters for the remaining models.

\textbf{Datasets}. To evaluate model performance on datasets in the size range of \datasetname, we use the three medical imaging datasets: CTSpine1K \cite{deng_CTSpine1K}, LiTS \cite{bilic_LITS}, LIDC-IDRI \cite{Armato_LIDC_IDRI}, in addition to \datasetname. These datasets include clinical CT scans with an in-plane resolution of $512 \times 512$, and a varying number of slices of human spine, liver, and thorax regions, respectively. The medical datasets are only available at one resolution, so models are trained on LR volumes downsampled from the HR volumes. For \datasetname, we consider two SR tasks: \datasetname (downsampled) tests the effect of downsampling, while \datasetname (registered) tests the effect of using scanned LR-HR paired data. For \datasetname (registered) at scale $2\times$, we downsample HR scans by a factor 2 to obtain $2\times$ resolution difference. \datasetname training and test splits are partly obtained by leaving out whole samples, and partly by separating single samples. For the vertebrae and femur samples of VoDaSuRe, we reserve whole scans for testing. For the remaining samples, we split the volumes along the axial direction to create volumes for training and testing, reserving $ \sim1/10^{\text{th}}$ of each volume for testing. See supplementary material for additional details. 

\textbf{Training}. All models are trained on a single H100 80GB GPU for 100K iterations using the AdamW optimizer \citep{Kingma_ADAM} with $\beta_{1} = 0.9$ and $\beta_{2} = 0.999$. We first train all models from scratch at $2\times$ upscaling using a batch size of $16$. For $4\times$ upscaling, we finetune models trained for $2\times$ upscaling for another 100K iterations using a batch size of $8$. The LR patch size is set to $32^3$ for all methods, or $32^2$ for 2D methods.  For augmentations, we use a combination of random 3D angular rotations and flipping in $(x,y,z)$, random contrast adjustment, and scaling. All model parameters are optimized using a constant learning rate and pure $L_{1}$ loss. 

\textbf{Performance evaluation}. We produce SR reconstructions of all volumes in the test set by tiled aggregation of SR predictions with an overlap of four voxels on all sides. Overlapping prediction regions are smoothed using a 3D Hanning window. 
Performance metrics Peak Signal-to-Noise Ratio (PSNR), Structural Similarity Index Measure (SSIM), Normalized Root Mean Square Error (NRMSE) and Learned Perceptual Image Patch Similarity (LPIPS) \cite{LPIPS} are computed slice-wise across the prediction volume and averaged over all non-zero slices. 
For 2D methods, metrics are evaluated on every $s^{\text{th}}$ slice, where $s$ is the SR scale. 

\subsection{In-domain experiments}
\label{subsec:in_domain_experiments}

\Cref{tab:table_quantitative_results_in_domain} shows a quantitative comparison of state-of-the-art SR methods on the evaluated datasets, with visual comparisons of all methods shown in \cref{fig:model_visualizations}. In medical imaging datasets CTSpine1K, LiTS and LIDC-IDRI, where LR is obtained by downsampling, all methods achieve impressive performance. Top-performing methods reach PSNR values of $\geq 40$~dB at scale $2\times$, and $\geq35$~dB at scale $4\times$. Conversely, in \datasetname (downsampled), all methods substantially drop in performance, despite using the same degradation method. The best PSNR is $25.50$~dB for $2\times$ and $19.08$~dB at scale $4\times$, with similar trends for SSIM, NRMSE and LPIPS. This shows that SR on \datasetname (downsampled) is substantially more challenging than on CTSpine1K, LiTS, and LICD-IDRI. Despite lower performance, all methods recover most of the structural variation.   

On \datasetname (registered), where we use scanned LR data, we observe a clear difference. All models produce noticeably more blurred output images compared to models trained on downsampled data, which is also reflected in lower performance. The best model achieves PSNR of $18.25$~dB at $2\times$ scale difference and $16.24$~dB at $4\times$, with similar trends in SSIM, NRMSE and LPIPS (see \cref{tab:table_quantitative_results_in_domain}). These differences show that models trained using pixel-wise loss fail to capture missing microstructural details in real LR data, and instead predict unrealistically smoothed outputs.

To assess the loss of high-frequency information in SR predictions, we compute the total variation (TV) for high-resolution images, predictions from synthetically downsampled inputs, and predictions from scanned LR data, see \cref{fig:TV_visualizations}. TV decreases notably in SR predictions using downsampled LR inputs compared with HR data, which can be explained by a general smoothing effect of SR. Yet, we observe an even lower TV in predictions using scanned LR data. This substantial loss of fine-scale structural detail confirms that SR of actual scanned LR data is a significantly harder SR task than upscaling downsampled LR volumes.

\subsection{Cross-domain experiments}
\label{subsec:in_domain_experiments}

To illustrate the difference between training on downsampled and real LR data, we perform cross-domain experiments on \datasetname, see \cref{tab:cross_domain_experiments} and \cref{fig:ablation_input}. We observe that models trained on downsampled LR data exhibit clear performance drops when evaluated using registered LR scans. This illustrates that training on downsampled data is a practice that cannot be successfully adapted to real data.

\subsection{Ablations}

\begin{figure}[t]
\centering
\renewcommand{\arraystretch}{1.0}

\resizebox{1.0\linewidth}{!}{%
\begin{minipage}{\linewidth}
\centering

\begin{subfigure}{\linewidth}
    \includegraphics[width=\linewidth]{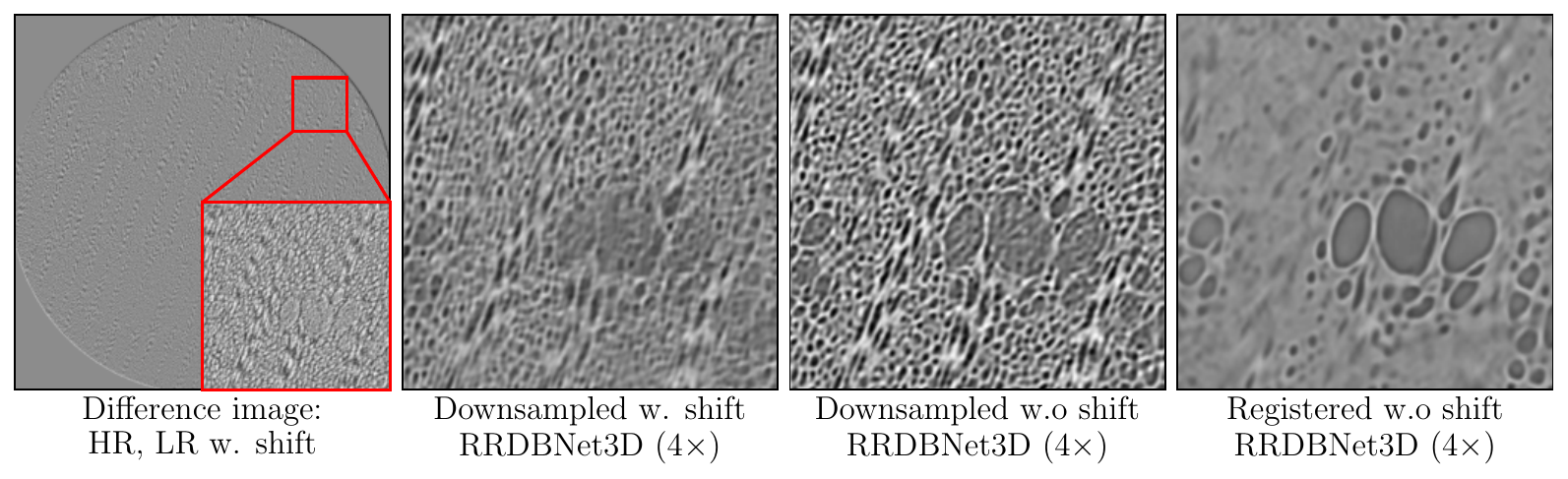}
    \subcaption{Ablation on the effect misregistration between LR and HR data.}
    \label{subfig:ablation_registration}
\end{subfigure}

\vspace{0.1cm}

\begin{subfigure}{\linewidth}
    \includegraphics[width=\linewidth]{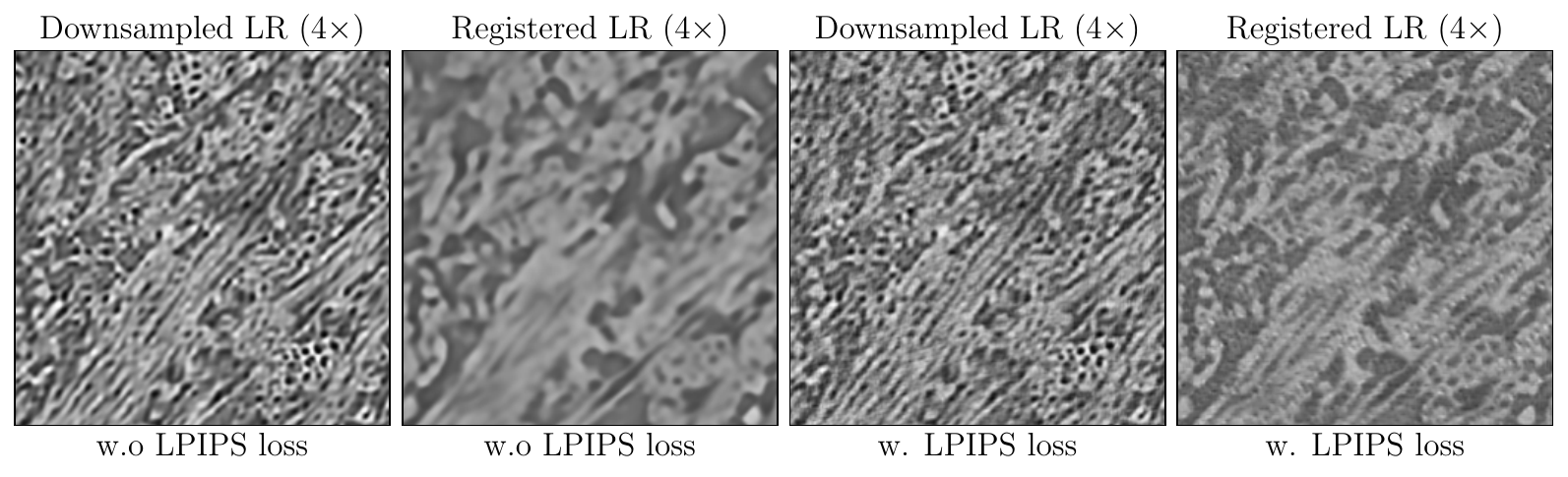}
    \subcaption{Ablation on the effect of perceptual (LPIPS) loss at scale $4\times$.}
    \label{subfig:ablation_LPIPS}
\end{subfigure}

\vspace{0.1cm}

\begin{subtable}{\linewidth}
\centering
\resizebox{\linewidth}{!}{%
\begin{NiceTabular}{cccc}[]
\toprule
\Block{1-1}{Method} & \Block{1-1}{Sample} & \Block{1-1}{Downsampling HR \& LR} & \Block{1-1}{\small{PSNR / SSIM / NRMSE / LPIPS}} \\ \midrule
MTVNet & Elm & original $(4\times)$ & 13.86 / .3296 / .4279 / .4219 \\
MTVNet & Elm & w. downsampling $(4\times)$ & 14.89 / .2588 / .3818 / .7277 \\
RRDBNet3D & Elm & original $(4\times)$ & 13.88 / .3440 / .4220 / .4122 \\
RRDBNet3D & Elm & w. downsampling $(4\times)$ & 14.84 / .2718 / .4183 / .6949 \\
\bottomrule
\end{NiceTabular}
}
\end{subtable}

\vspace{0.1cm}

\begin{subfigure}{\linewidth}
    \includegraphics[width=\linewidth]{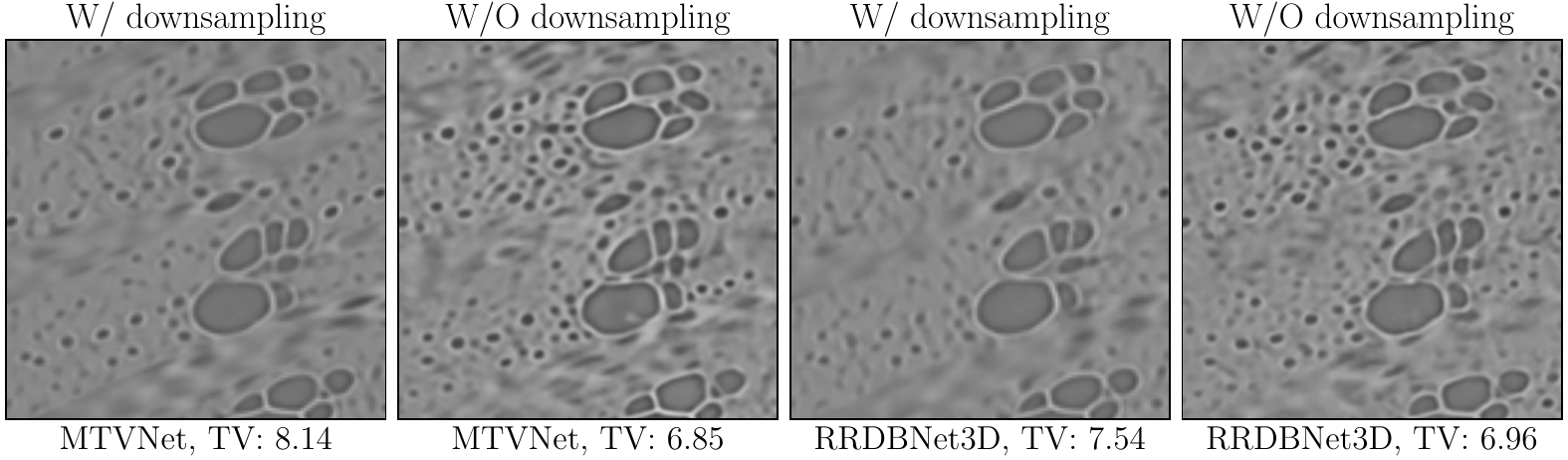}
    \subcaption{Ablation on the effect of downsampling both HR and real LR data.}
    \label{subfig:downsampling}
\end{subfigure}

\vspace{0.1cm}

\begin{subtable}{\linewidth}
\centering
\resizebox{\linewidth}{!}{%
\begin{NiceTabular}{cccc}[]
\toprule
\Block{1-1}{Method} & \Block{1-1}{Sample} & \Block{1-1}{Include during training} & \Block{1-1}{\small{PSNR / SSIM / NRMSE / LPIPS}} \\ \midrule
MTVNet & Wood & \cmark & 16.13 / .4942 / .3489 / .3933 \\
MTVNet & Wood & \xmark & 15.83 / .4953 / .3614 / .4264 \\
RRDBNet3D & Wood & \cmark & 16.16 / .5123 / .3447 / .3818 \\
RRDBNet3D & Wood & \xmark & 15.86 / .4910 / .3603 / .4526 \\
\bottomrule
\end{NiceTabular}
}
\end{subtable}

\vspace{0.1cm}

\begin{subfigure}{\linewidth}
    \includegraphics[width=\linewidth]{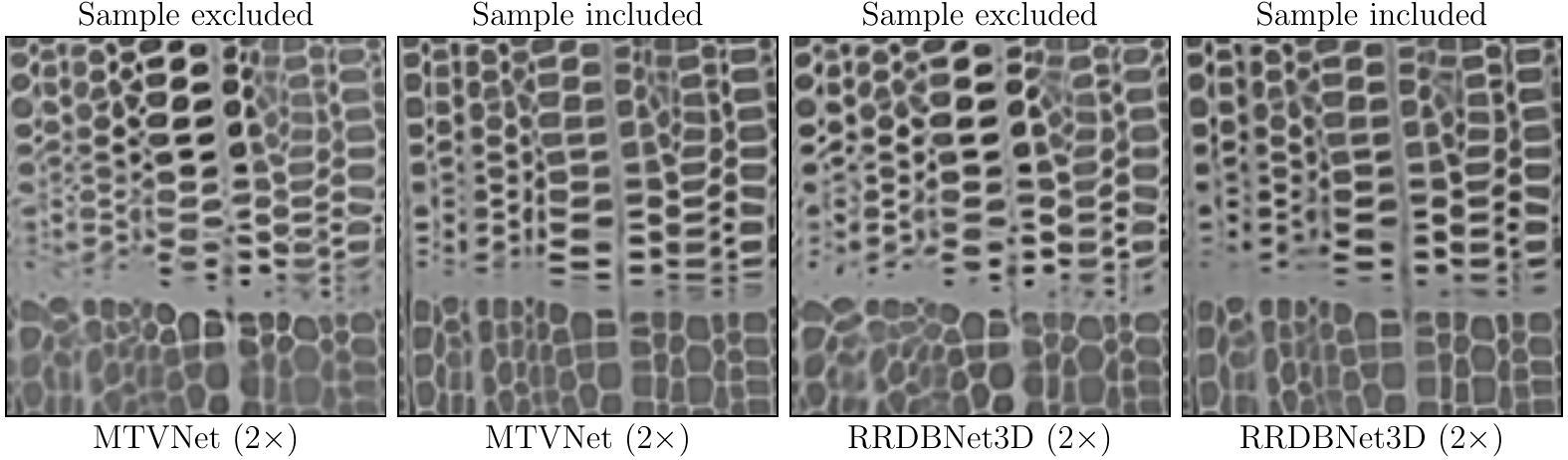}
    \subcaption{Ablation on the effect of excluding specific samples during training.}
    \label{subfig:ablation_samples}
\end{subfigure}

\end{minipage}%
} 

\vspace{-0.1cm}

\caption{Ablation experiments on \datasetname. }
\label{tab:ablation_study}
\end{figure}

We conduct four ablation experiments using \datasetname, see \cref{tab:ablation_study}. First, we test whether the observed smoothing is inducible via registration error. We purposely misalign the downsampled Elm volume and retrain RRDBNet3D, see \cref{subfig:ablation_registration}. While misalignment reduces prediction sharpness, it does not reproduce the characteristic smoothing observed when training on real LR scans, suggesting that acquisition effects are the primary cause of the observed domain shift. 

Next, we investigate the use of perceptual loss for bridging the domain gap, see \cref{subfig:ablation_LPIPS}. Training RRDBNet3D using a combination of L1 and LPIPS loss $(\lambda =0.02)$ results in noticable increase in texture, yet predictions remain unrealistic compared to HR data, and the domain gap persists.

Third, we test the effect of creating $2\times$ scale difference in scanned data by downsampling the HR scans, see \cref{subfig:downsampling}. By also downsampling the LR scans $2\times$, we obtain a scale difference of $4\times$. Training RRDBNet3D and MTVNet with and without downsampling, we observe minor performance differences and similar smoothing effects, indicating that the domain shift is intrinsic to real LR data. This confirms the $2\times$ downsampling of the HR scans as a viable practice.


Finally, we test generalization to unseen materials by training on bamboo, oak, and larch wood while evaluating on elm and cypress, see \cref{subfig:ablation_samples}. We obtain comparable results across all metrics, demonstrating that models trained on \datasetname enable generalization across similar microstructures, albeit with similarly smoothed predictions.

\section{Discussion and conclusion}
\label{sec:conclusion}

We introduce \datasetname, a high-resolution benchmark for volumetric super-resolution featuring both synthetically downsampled and physically acquired low-resolution scans obtained using identical imaging setups. \datasetname includes diverse microstructures from biological and non-biological materials, including wood, composite materials, and bone, filling a gap between scientific and biomedical imaging and enabling SR benchmarking across complex 3D structures. To our knowledge, \datasetname is the largest volumetric dataset with multi-resolution data for all scans.

Training state-of-the-art SR models using pixel-wise loss on real LR data, we reveal a clear domain shift in SR predictions. All models produce spatially averaged predictions lacking the high-frequency information observed in both HR and LR data. This effect is not observable in downsampled data, suggesting that SR models primarily learn to invert degradation instead of reconstructing microstructures. Using total variation, we confirm a loss of high-frequency spatial detail in SR predictions, supported by consistent drops in performance metrics and visual observations. This effect is not reproducible via misregistration alone, indicating that the domain gap arises from the LR acquisition, and while adding perceptual loss produces more textured predictions, it does not resolve the domain gap. These findings highlight the need for more advanced approaches that can address the domain shift observed in real data to retain the relevance of SR for scientific and practical applications.

Using Local Attribution Mapping \cite{gu2021interpreting}, we analyzed how SR models leverage spatial context across datasets. All models exhibited higher diffusion indices (DI) on \datasetname, particularly ViT-based methods (HAT, SuperFormer, MTVNet), indicating greater reliance on image context. However, we observed no clear correlation between DI and performance, suggesting that contextual dependency alone does not guarantee SR accuracy. See Supp. for details.

In summary, \datasetname provides a challenging benchmark for studying SR generalization under realistic acquisition conditions, supporting the development of more robust and physically grounded volumetric SR models.

\newpage

\section*{Acknowledgments}
\label{sec:acknowledgments}
Research reported in this publication is supported by the Infrastructure for Quantitative AI-based Tomography (QUAITOM) supported by the Novo Nordisk Foundation (Grant number NNF21OC0069766) and the Multiscale label-free 3D x-ray imaging: Visualizing cells and tissue architecture simultaneously (Xtreme-CT) supported by the Novo Nordisk Foundation (Grant number NNF22OC0077698).

{
    \small
    \bibliographystyle{ieeenat_fullname}
    \bibliography{main, references}
}

\clearpage
\maketitlesupplementary

\section{\datasetname dataset overview}

A detailed overview of all 16 samples in the \datasetname dataset is provided in \cref{tab:table_sample_overview_V2}, including volume shapes, slice splits (when applicable), voxel sizes, and scanning devices. The table lists only the physically acquired scans (HR/LR) and the registered LR volumes. Additional downsampled pyramid levels produced during OME-Zarr conversion are omitted for clarity. Note that the voxel sizes of the LR, and registered LR scans differ slightly, as the voxel size of the acquired LR scans did not exactly match the desired $4\times$ resolution difference compared with HR. This discrepancy is accounted for during the registration procedure. 

\textbf{Sample selection}. To ensure a diverse set of structural characteristics, we intentionally include materials with varying degrees of microstructural complexity. We chose wood samples due to their well-organized tubular structures, as well as MDF and cardboard for their more chaotic arrangements of layers and fibers. We also chose to incorporate bone samples (femur, vertebrae, and animal bone) to have volumes with smoother structures typically seen in medical imaging datasets of clinical volumes. The finest microstructures in wood, MDF, and cardboard samples lie near the resolution limit of the LR scans but are clearly visible in the HR scans. This design choice ensures meaningful super-resolution scenarios where relevant structural details are partially lost in the LR input.

\textbf{Stitching \& reconstruction}. All scans are reconstructed using the standard software provided with each scanner. Similarly, stitching of multiple vertical scans is performed using the native stitching tools of the respective devices.

\section{\datasetname preprocessing}
\label{sec:vodasure_data_analysis_pipeline}

\begin{figure}[t]
   \centering
   \includegraphics[width=1.00\linewidth]{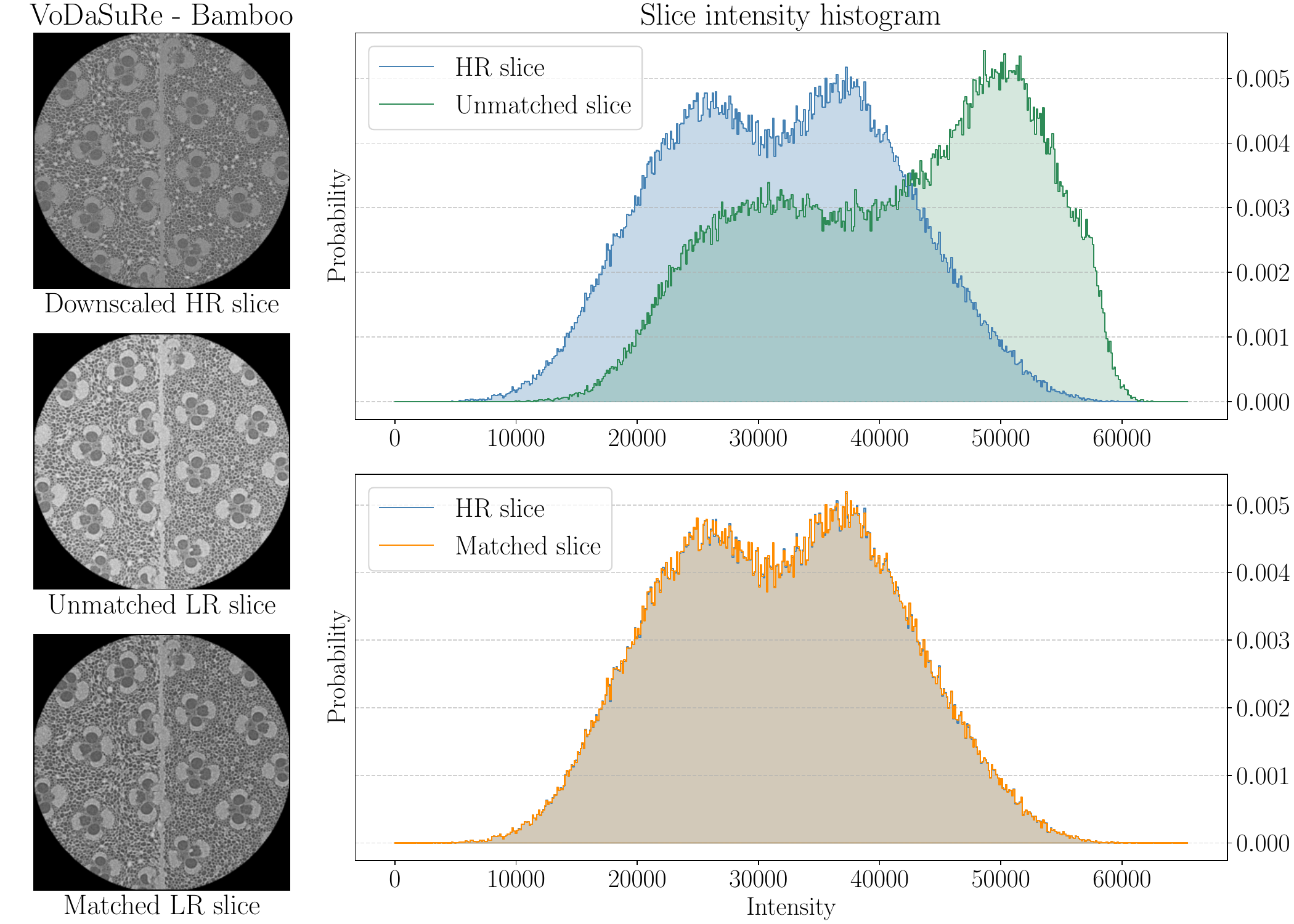}
   \caption{Visualization of the intensity matching procedure used in \datasetname. The intensity distribution of registered LR slices is adjusted to match the distribution of downsampled HR slices. }
   \label{fig:supp_matching}
\end{figure}

\textbf{Intensity matching}. During the curation of \datasetname, we observed notable differences in intensity distributions between all acquired LR and HR scans. For LR scans, the reduced cone-beam dispersion of the CT setup resulted in increased detector counts, improved signal-to-noise ratio, and higher contrast compared with HR acquisition. In some scans, the effect of region-of-interest scanning in high resolution (the scan region surrounded by material that is not accounted for in the reconstruction) resulted in small intensity differences between HR and LR scans in regions furthest from the rotational axis. To mitigate this, we applied intensity matching of registered LR slices to downsampled HR slices. \cref{fig:supp_matching} shows the effect of intensity matching using bamboo. The LR slice appears noticeably brighter with stronger contrast than the HR slice, which is also reflected in the intensity histograms of the two slices. After intensity matching, the intensity profile of the LR slice matches that of the HR slice but retains the same structural information. 

We initially attempted to match the intensities of registered LR slices directly to HR slices, but found that this led to unrealistic intensity scaling. The HR slices contain a significantly larger proportion of high-intensity voxels due to their higher resolution, whereas these details are spatially averaged in the LR scans. Consequently, direct HR–LR matching causes the LR slices to become oversaturated. To avoid this, we first downsample the HR slices to the LR voxel size and then perform intensity matching. This downsampling suppresses high-frequency content while maintaining overall intensity statistics, resulting in more stable and physically meaningful intensity alignment.

\begin{figure*}[t]
  \centering
  \begin{subfigure}{0.96\linewidth}
    \includegraphics[width=1.0\linewidth]{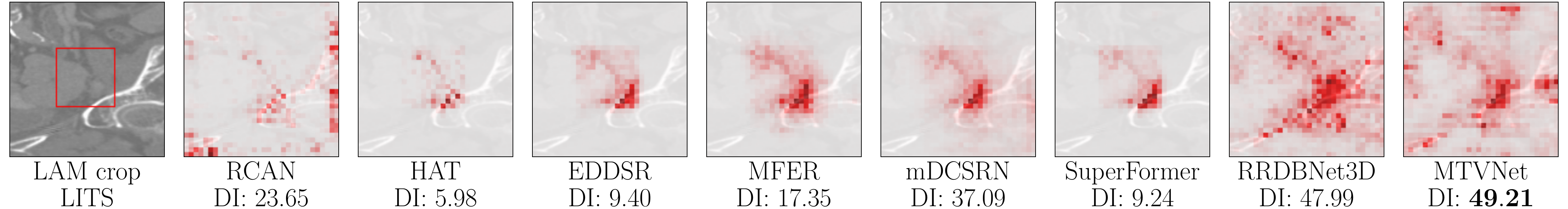}
  \end{subfigure}
  \hfill
  \begin{subfigure}{0.96\linewidth}
    \includegraphics[width=1.0\linewidth]{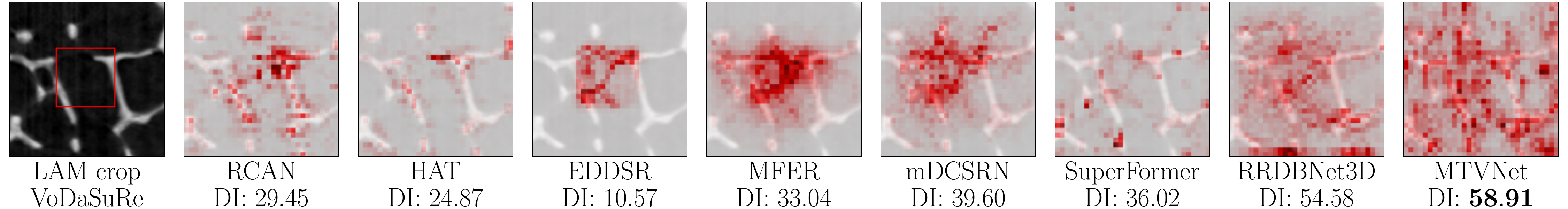}
  \end{subfigure}
  \hfill
  \begin{subfigure}{0.96\linewidth}
    \includegraphics[width=1.0\linewidth]{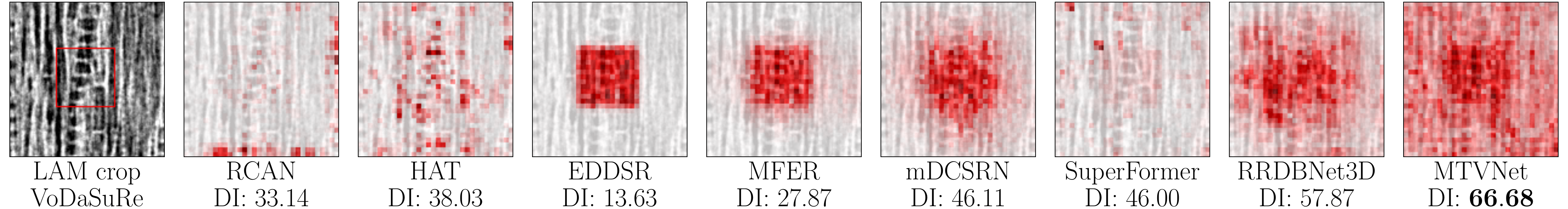}
  \end{subfigure}
  
  \caption{LAM comparisons of SR models. Top row: example from CTSpine1K, middle and bottom row: examples from \datasetname. The highest DI $\uparrow$ is highlighted in \textbf{bold}.}
  \label{fig:LAM_visualizations}
\end{figure*}

\begin{figure}[t]
   \centering
   \includegraphics[width=1.00\linewidth]{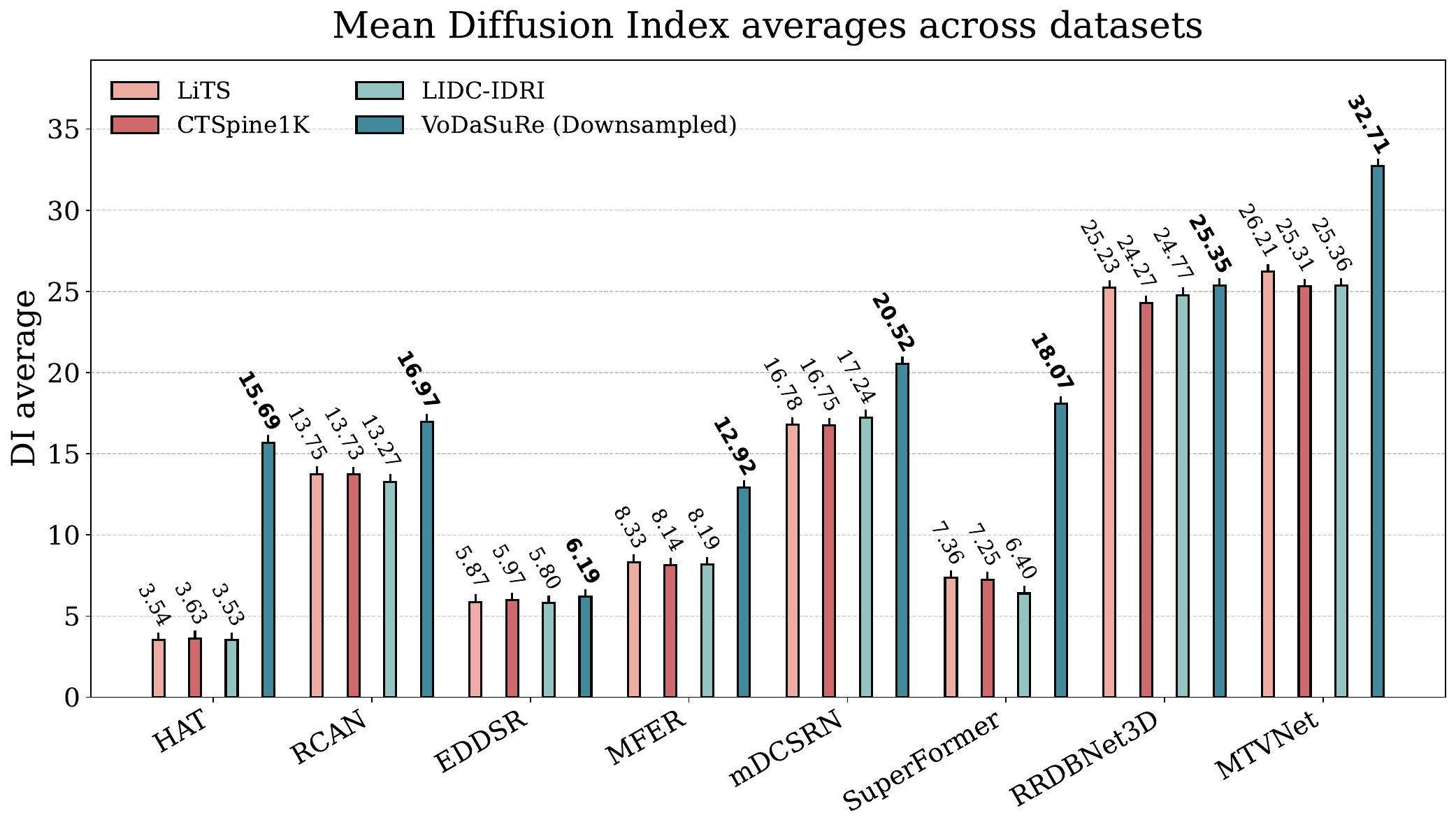}
   \caption{Diffusion index (DI) averages using datasets CTSpine1K, LiTS and LIDC-IDRI for all SR models. The highest DI $\uparrow$ scores for each dataset are highlighted in \textbf{bold}.}
   \label{fig:diffusion_index_averages}
\end{figure}

\begin{figure*}[t]
  \includegraphics[width=1.00\linewidth]{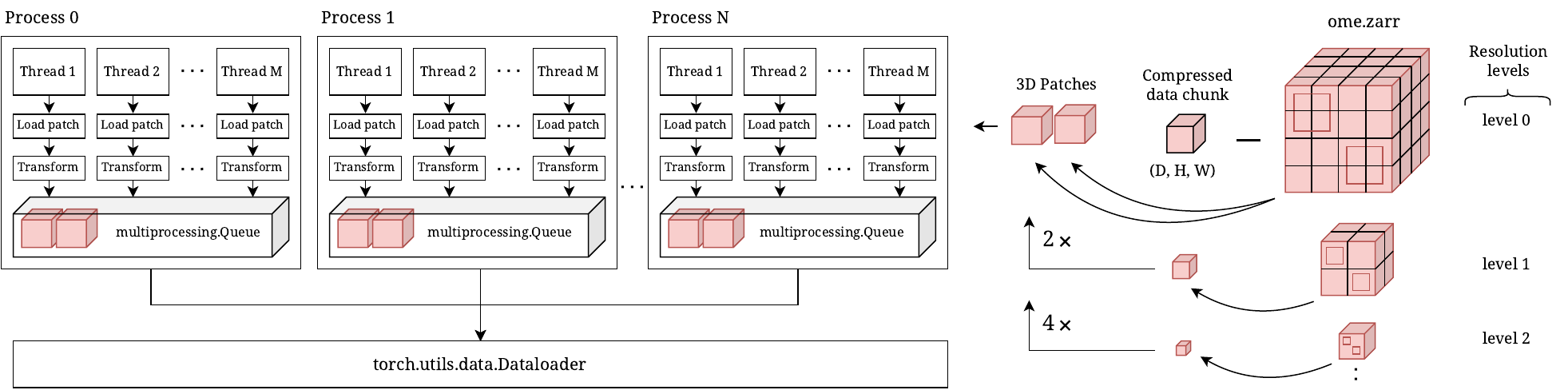}
    \caption{Illustration of the data loading pipeline for \datasetname based on the OME-Zarr data format.} 
  \label{fig:supp_dataloader_overview}
\end{figure*}

\begin{figure*}[t]
  \centering
  \begin{subfigure}{1.00\linewidth}
    \includegraphics[width=1.0\linewidth]{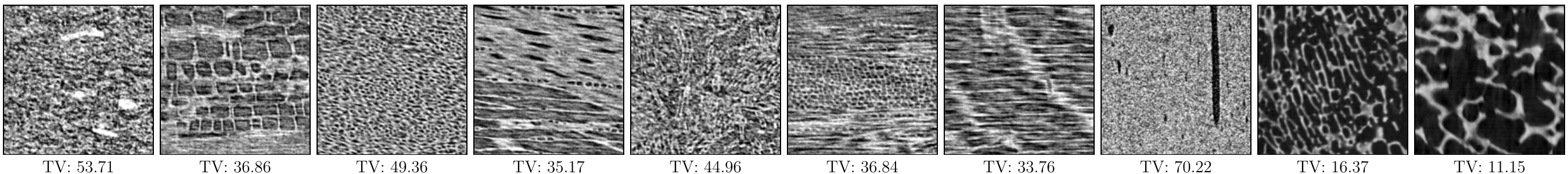}
  \end{subfigure}
  \hfill
  \begin{subfigure}{1.00\linewidth}
    \includegraphics[width=1.0\linewidth]{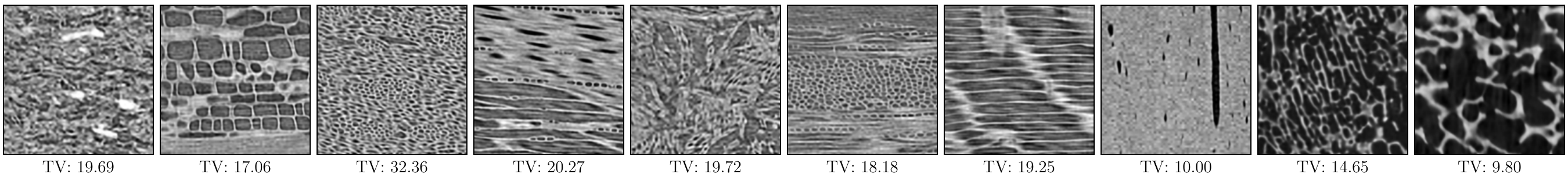}
  \end{subfigure}
   \hfill
  \begin{subfigure}{1.00\linewidth}
    \includegraphics[width=1.0\linewidth]{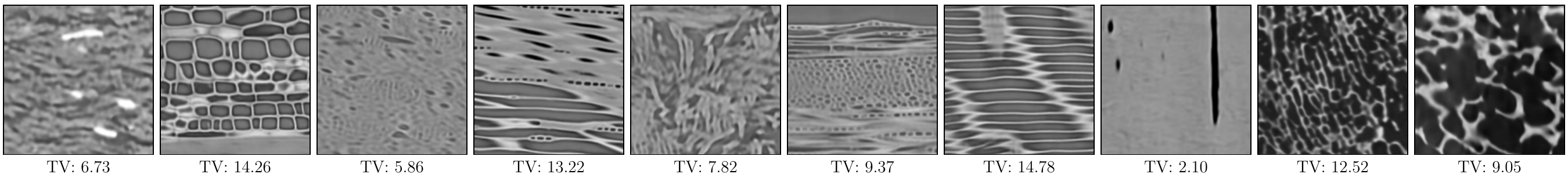}
  \end{subfigure}
  \hfill
  \begin{subfigure}{1.00\linewidth}
    \includegraphics[width=1.0\linewidth]{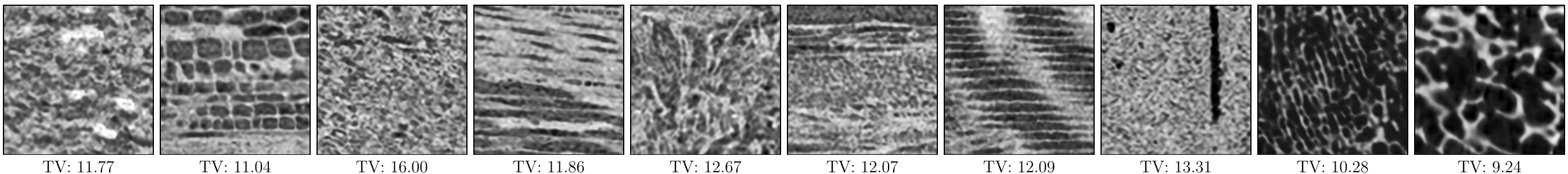}
  \end{subfigure}
  
  \caption{Visualizations from \datasetname. From top to bottom: HR data, model predictions using downsampled LR data, model predictions using real LR data, and model predictions obtained by training on downsampled data but evaluating using real LR data input. All outputs are obtained at $4\times$ upscaling using RRDBNet3D. Total variation (TV) is shown for each slice.}
  \label{fig:supp_TV_visualizations}
\end{figure*}

\begin{figure*}[t]
  \centering
  \begin{subfigure}{1.00\linewidth}
    \includegraphics[width=1.0\linewidth]{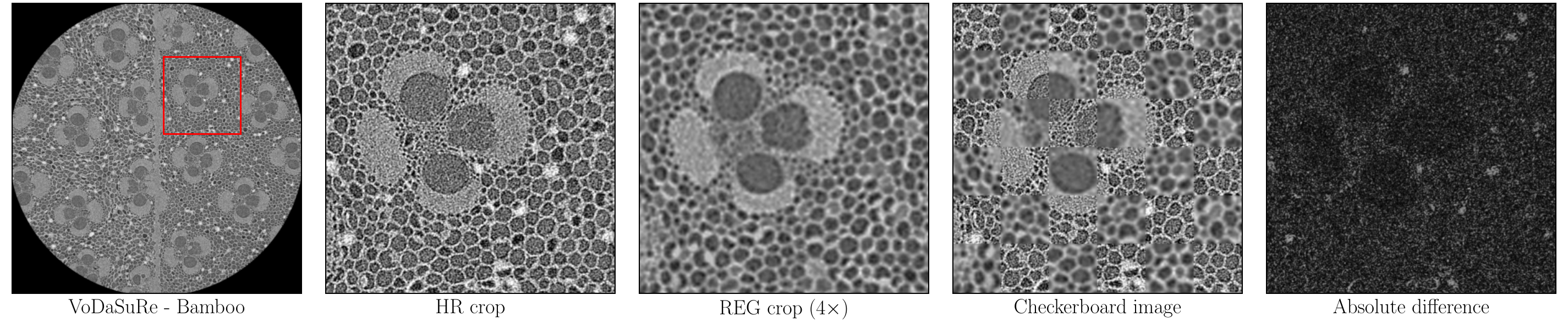}
  \end{subfigure}
  \hfill
  \begin{subfigure}{1.00\linewidth}
    \includegraphics[width=1.0\linewidth]{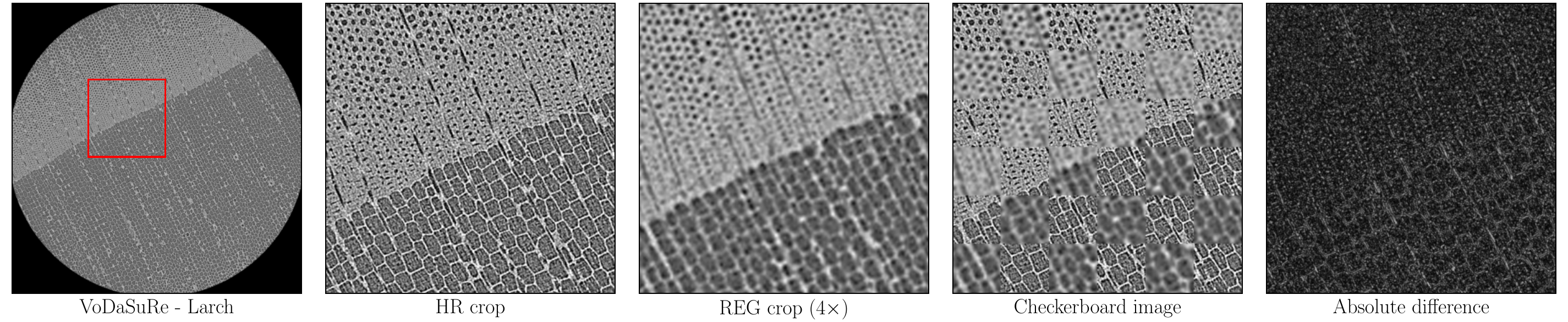}
  \end{subfigure}
  \hfill
  \begin{subfigure}{1.00\linewidth}
    \includegraphics[width=1.0\linewidth]{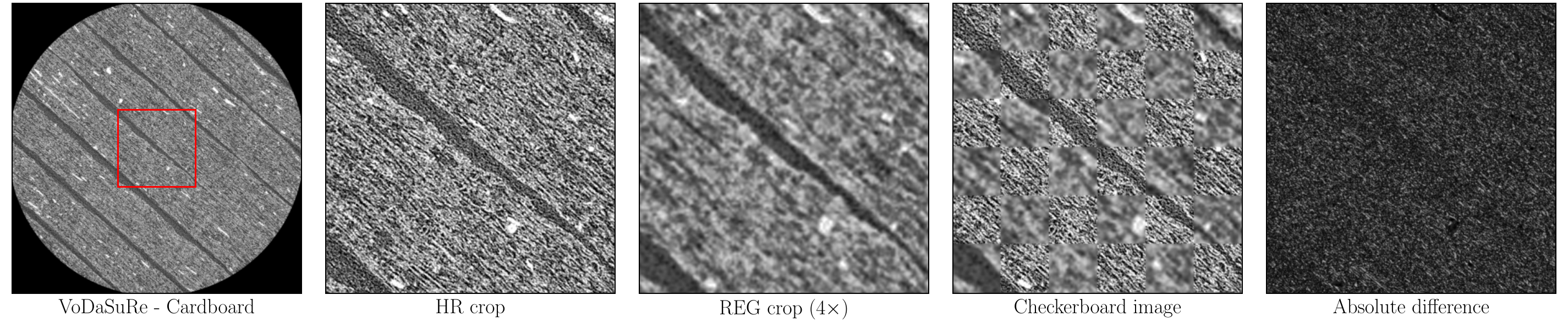}
  \end{subfigure}
  \hfill
  \begin{subfigure}{1.00\linewidth}
    \includegraphics[width=1.0\linewidth]{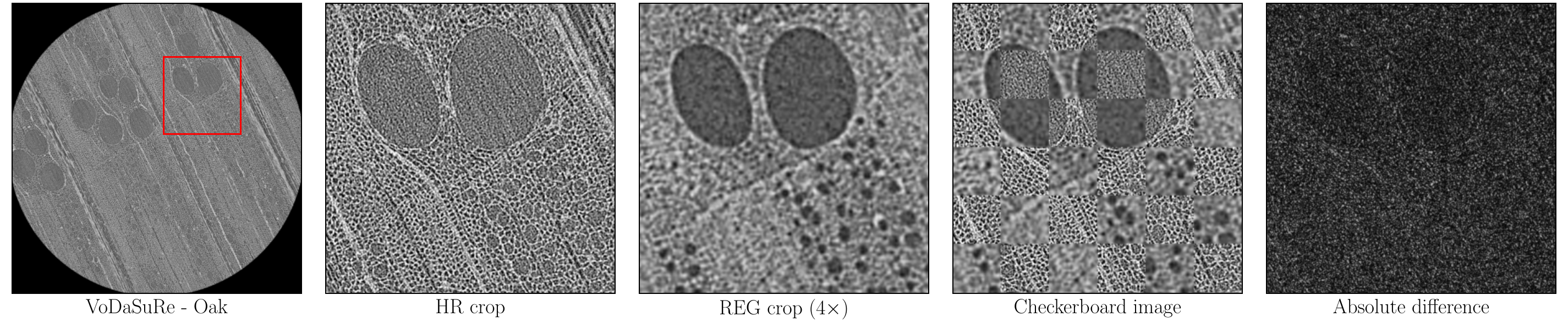}
  \end{subfigure}
  \hfill
  \begin{subfigure}{1.00\linewidth}
    \includegraphics[width=1.0\linewidth]{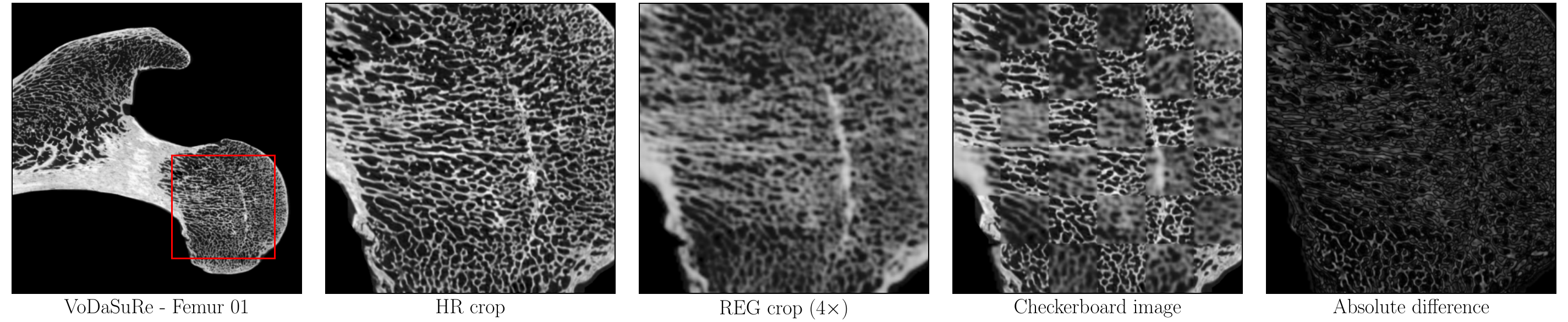}
  \end{subfigure}
  \hfill
  \begin{subfigure}{1.00\linewidth}
    \includegraphics[width=1.0\linewidth]{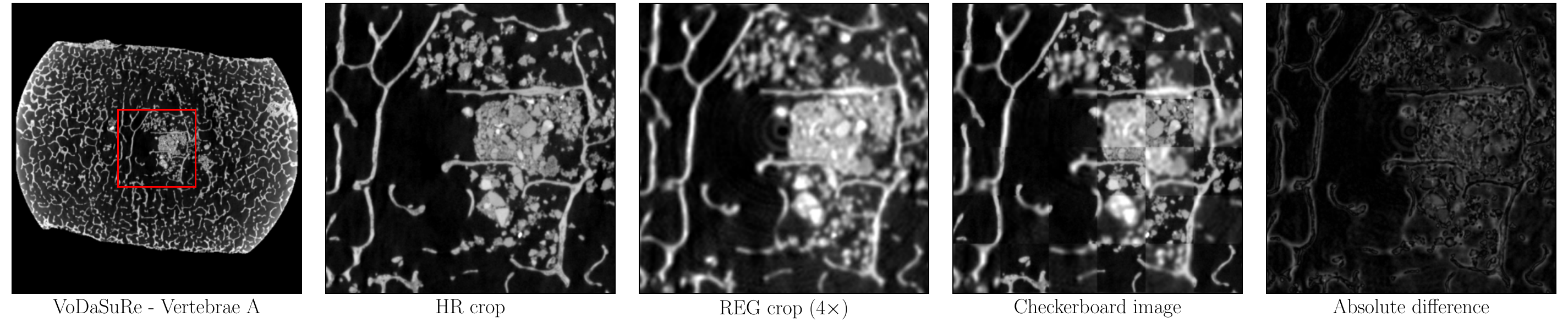}
  \end{subfigure}
  
  \caption{Evaluation of HR-LR registrations in \datasetname. From left to right: Full HR slice, cropped HR slice, cropped registered LR slice, checkerboard image, and absolute difference image between HR and interpolated LR slice.}
  \label{fig:supp_registration}
\end{figure*}

\textbf{Registration}. \cref{fig:supp_registration} illustrates the accuracy of the HR-LR registration after intensity matching. Cropped regions from corresponding HR and registered LR volumes highlight the expected loss in microstructural detail. To assess spatial alignment, we create checkerboard visualizations and absolute difference images. The checkerboard images confirm the continuity of structures across the HR and registered LR volumes and demonstrates the effectiveness of our registration procedure. Similarly, the absolute difference images of the HR, and bicubic-interpolated LR slices validate the alignment, and also reveal the high-frequency information absent in the LR volumes.

\begin{table*}[t]
\centering
\resizebox{\textwidth}{!}{

\renewcommand{\arraystretch}{1.1}
\begin{scriptsize}
\begin{NiceTabularX}{\textwidth}{
    X
    l
    X   
    c
    c
    c
    c
}[hvlines]
Sample name & Scan & Volume shape $(\text{D}\times \text{H}\times \text{W})$ & Slice split (train/test) & Voxel size $\left[\mu\text{m}\right]$ & Scanning device & Data size \\

\Block{3-1}{Bamboo} & High-resolution & \begin{scriptsize} $5440 \times 1920 \times 1920$\end{scriptsize} & 4960 / 480 & $1.671 \times 1.671 \times 1.671$ & \Block{3-1}{Zeiss Versa 520} & $37.4$ GB \\

& Low-resolution &  $3520 \times 1920 \times 1920$   & -  & $6.637 \times 6.637  \times 6.637$ & & $24.2$ GB \\

& Registered & $1360 \times 480 \times 480$   &  1240 / 120  & $6.684 \times 6.684  \times 6.684$ &  & $597.7$ MB \\

\Block{3-1}{Cardboard} & High-resolution & \begin{scriptsize} $5120 \times 1920 \times 1920$\end{scriptsize} & 4640 / 480 & $2.031 \times 2.031 \times 2.031$ & \Block{3-1}{Zeiss Versa 520} & $35.2$ GB \\

& Low-resolution & $3360 \times 1920 \times 1920$   & -  & $8.017 \times 8.017  \times 8.017$ & & $23.1$ GB \\

& Registered &  $1280 \times 480 \times 480$  &  1160 / 120  & $8.124 \times 8.124  \times 8.124$ &  & $562.5$ MB \\

\Block{3-1}{Cypress} & High-resolution & \begin{scriptsize} $5440 \times 1920 \times 1920$\end{scriptsize} & 4960 / 480 & $1.671 \times 1.671 \times 1.671$ & \Block{3-1}{Zeiss Versa 520} & $37.4$ GB \\

& Low-resolution & $1920 \times 1920 \times 1920$   & -  & $6.636 \times 6.636  \times 6.636$ & & $13.2$ GB \\

& Registered &  $1360 \times 480 \times 480$  &  1240 / 120  & $6.684 \times 6.684  \times 6.684$ &  & $597.7$ MB \\

\Block{3-1}{Elm} & High-resolution & \begin{scriptsize} $5440 \times 1920 \times 1920$\end{scriptsize} & 4960 / 480 & $1.671 \times 1.671 \times 1.671$ & \Block{3-1}{Zeiss Versa 520} & $37.4$ GB \\

& Low-resolution & $3520  \times 1920 \times 1920$   & -  & $6.637 \times 6.637  \times 6.637$ & & $24.2$ GB \\

& Registered &  $1360 \times 480 \times 480$ &  1240 / 120  & $6.684 \times 6.684  \times 6.684$ &  & $597.7$ MB \\

\Block{3-1}{MDF} & High-resolution & \begin{scriptsize} $3680 \times 1920 \times 1920$\end{scriptsize} & 3200 / 480 & $1.671 \times 1.671 \times 1.671$ & \Block{3-1}{Zeiss Versa 520} & $25.3$ GB \\

& Low-resolution & $3520  \times 1920 \times 1920$   & -  & $6.637 \times 6.637  \times 6.637$ & & $24.2$ GB \\

& Registered &  $920 \times 480 \times 480$  &  800 / 120  & $6.685 \times 6.685  \times 6.685$ &  & $404.3$ MB \\

\Block{3-1}{Ox bone} & High-resolution & \begin{scriptsize} $4960 \times 1920 \times 1920$\end{scriptsize} & 4480 / 480 & $1.199 \times 1.199 \times 1.199$ & \Block{3-1}{Zeiss Versa 520} & $34.1$ GB \\

& Low-resolution & $1920 \times 1920 \times 1920$   & -  & $ 4.798  \times 4.798  \times  4.798 $ & & $13.2$ GB \\

& Registered &  $1240 \times 480 \times 480$  &  1120 / 120  & $4.796 \times 4.796  \times 4.796$ &  & $544.9$ MB \\

\Block{3-1}{Oak} & High-resolution & \begin{scriptsize} $5440 \times 1920 \times 1920$\end{scriptsize} & 4960 / 480 & $1.671 \times 1.671 \times 1.671$ & \Block{3-1}{Zeiss Versa 520} & $37.4$ GB \\

& Low-resolution & $3200 \times 1920 \times 1920$   & -  & $6.637 \times 6.637  \times 6.637$ & & $22.0$ GB \\

& Registered &  $1360 \times 480 \times 480$  &  1240 / 120  & $6.684 \times 6.684  \times 6.684$ &  & $597.7$ MB \\

\Block{3-1}{Larch} & High-resolution & \begin{scriptsize} $5120 \times 1920 \times 1920$\end{scriptsize} & 4640 / 480 & $1.669 \times 1.669 \times 1.669$ & \Block{3-1}{Zeiss Versa 520} & $35.2$ GB \\

& Low-resolution & $3200 \times 1920 \times 1920$   & -  & $6.637 \times 6.637  \times 6.637$ & & $22.0$ GB \\

& Registered & $1280 \times 480 \times 480$   &  1160 / 120  & $6.674 \times 6.674  \times 6.674$ &  & $562.5$ MB \\

\Block{3-1}{Femur 15} & High-resolution & \begin{scriptsize} $1600 \times 1280 \times 1920$\end{scriptsize} & \Block{3-1}{Train} & $58 \times 58 \times 58$ & \Block{3-1}{Nikon XT H 225} & $7.3$ GB \\

& Low-resolution & $600 \times 600 \times 600$   &   & $ 232  \times 232  \times 232 $ & & $412.0$ MB \\

& Registered &  $400 \times 320 \times 480$  &    & $ 232  \times 232  \times 232 $ &  & $117.2$ MB \\

\Block{3-1}{Femur 21} & High-resolution & \begin{scriptsize} $1280 \times 1600 \times 1760$\end{scriptsize} & \Block{3-1}{Train} & $58 \times 58 \times 58$ & \Block{3-1}{Nikon XT H 225} & $6.7$ GB \\

& Low-resolution & $600 \times 600 \times 600$   &   & $ 232  \times 232  \times 232 $ & & $412.0$ MB \\

& Registered &  $320 \times 400 \times 440$  &    & $ 232  \times 232  \times 232 $ &  & $107.4$ MB \\

\Block{3-1}{Femur 74} & High-resolution & \begin{scriptsize} $1120 \times 1760 \times 1600$\end{scriptsize} & \Block{3-1}{Train} & $58 \times 58 \times 58$ & \Block{3-1}{Nikon XT H 225} & $5.9$ GB \\

& Low-resolution & $600 \times 600 \times 600$   &   & $ 232  \times 232  \times 232 $ & & $412.0$ MB \\

& Registered &  $280 \times 440 \times 400$  &    & $ 232  \times 232  \times 232 $ &  & $94.0$ MB \\

\Block{3-1}{Femur 01} & High-resolution & \begin{scriptsize} $960 \times 1440 \times 1600$\end{scriptsize} & \Block{3-1}{Test} & $58 \times 58 \times 58$ & \Block{3-1}{Nikon XT H 225} & $4.1$ GB \\

& Low-resolution & $600 \times 600 \times 600$   &   & $ 232  \times 232  \times 232 $ & & $412.0$ MB \\

& Registered &  $240 \times 360 \times 400$  &    & $ 232  \times 232  \times 232 $ &  & $65.9$ MB \\

\Block{3-1}{Vertebrae A} & High-resolution & \begin{scriptsize} $1920 \times 1920 \times 1920$\end{scriptsize} & \Block{3-1}{Train} & $22 \times 22 \times 22$ & \Block{3-1}{Nikon XT H 225} & $13.2$ GB \\

& Low-resolution & $800 \times 960 \times 640$   &   & $88 \times 88 \times 88$ & & $937.5$ MB \\

& Registered &  $480 \times 480 \times 480$  &    & $88 \times 88 \times 88$ &  & $210.9$ MB \\

\Block{3-1}{Vertebrae B} & High-resolution & \begin{scriptsize} $1920 \times 1920 \times 1920$\end{scriptsize} & \Block{3-1}{Train} & $22 \times 22 \times 22$ & \Block{3-1}{Nikon XT H 225} & $13.2$ GB \\

& Low-resolution & $800 \times 960 \times 640$   &   & $88 \times 88 \times 88$ & & $937.5$ MB \\

& Registered &  $480 \times 480 \times 480$  &    & $88 \times 88 \times 88$ &  & $210.9$ MB \\

\Block{3-1}{Vertebrae C} & High-resolution & \begin{scriptsize} $1920 \times 1920 \times 1920$\end{scriptsize} & \Block{3-1}{Train} & $22 \times 22 \times 22$ & \Block{3-1}{Nikon XT H 225} & $13.2$ GB \\

& Low-resolution & $960 \times 800 \times 960$   &   & $88 \times 88 \times 88$ & & $1.4$ GB \\

& Registered &  $480 \times 480 \times 480$  &    & $88 \times 88 \times 88$ &  & $210.9$ MB \\

\Block{3-1}{Vertebrae D} & High-resolution & \begin{scriptsize} $1920 \times 1920 \times 1920$\end{scriptsize} & \Block{3-1}{Test} & $22 \times 22 \times 22$& \Block{3-1}{Nikon XT H 225} & $13.2$ GB \\

& Low-resolution & $960 \times 800 \times 960$    &   & $88 \times 88 \times 88$& & $1.4$ GB \\

& Registered & $480 \times 480 \times 480$  &    & $88 \times 88 \times 88$ &  & $210.9$ MB \\



\end{NiceTabularX}
\end{scriptsize}
} 

\caption{Overview of \datasetname, including sample names, volume shapes, slice splits for training and testing, voxel sizes and scanning devices. For vertebrae and femur samples, we reserve whole scans for training/test, while remaining scans are split into training/test slices.}
\vspace{-0.05in}
\label{tab:table_sample_overview_V2}
\end{table*}

\begin{table*}[t]
\centering
\renewcommand{\arraystretch}{1.0}

\resizebox{\textwidth}{!}{%
\begin{footnotesize}
\begin{NiceTabular}{c*{8}{c}}[]
\toprule
Method 
& \Block{1-1}{RCAN} & \Block{1-1}{HAT}
& \Block{1-1}{EDDSR} & \Block{1-1}{SuperFormer}
& \Block{1-1}{MFER} & \Block{1-1}{mDCSRN}
& \Block{1-1}{MTVNet} & \Block{1-1}{RRDBNet3D}
\\ \midrule

No. of parameters
&  15.6M & 20.8M
&  0.8M & 20.4M
&  1.7M & 1.7M
&  67.0M & 26.1M
\\ \midrule

Avg. training time
& 9.47 \text{h} & 5.64 \text{h}
&  8.09 \text{h} &  32.85 \text{h}
&  50.65 \text{h} & 7.76 \text{h}
& 31.05 \text{h} &  16.48 \text{h}
\\

\bottomrule
\end{NiceTabular}
\end{footnotesize}
}

\caption{Average training time of SR methods at $4\times$ upscaling. The measured times is the average time to complete 100K training iterations across datasets CTSpine1K, LiTS, LIDC-IDRI and VoDaSuRe.}
\vspace{-0.05in}
\label{tab:supp_training_duration}
\end{table*}

\section{LAM analysis}
\label{subsec:LAM_analysis}





To assess the degree of contextual dependency of SR predictions across datasets, we employ Local Attribution Mapping (LAM) \cite{gu2021interpreting}. Using LAM, we compare the spread of input voxel attributions for SR models trained on datasets with fine microstructures, e.g. \datasetname, and models trained on medical data with smoother variations. Fig. \ref{fig:LAM_visualizations} shows slice-averaged LAM results at scale $4\times$, where regions of higher intensities indicate stronger pixel/voxel contributions. We also report the slice-wise average Diffusion Index (DI) \citep{gu2021interpreting} as an estimate for overall context usage. Examples show that all models leverage broader involvement of input voxels in \datasetname. To quantify this effect, we evaluate all SR models on 100 randomly sampled 3D patches from CTSpine1K, LiTS, LIDC-IDRI, and \datasetname, and calculate the average DI of all models across all patches, see \cref{fig:diffusion_index_averages}. We observe consistently higher DI across all methods, meaning SR models rely on broader spatial context in \datasetname compared with CTSpine1K, LiTS and LIDC-IDRI. This suggests that long-range information is more important in \datasetname than in medical imaging datasets, where models rely more on local image context. In particular, we observe ViT-based methods HAT, SuperFormer and MTVNet exhibiting noticeably greater increases in diffusion index using \datasetname compared with CNN-based methods. Despite this, we did not find a correlation in performance, as the CNN-based RRDBNet3D was the overall strongest baseline in both medical datasets and \datasetname. 



\section{OME-Zarr dataloader}

\cref{fig:supp_dataloader_overview} shows our data loading pipeline. We instantiate $N$ worker processes that concurrently load volumetric patches from disk, with each worker using multiple threads that each maintain their own data queues to avoid contention. After loading and augmentation, patches are stored in the respective thread's data queue. During runtime, the main process collates batches of patches from all worker processes to maintain data throughput. This way, our pipeline scales to extremely large datasets, as full volumes are never held in system memory. Each OME-Zarr store in \datasetname contains multiple resolution levels. By sampling patches from corresponding regions at different levels, we conveniently generate LR–HR pairs. The resolution gap between pyramid levels defines the SR scale, with each step yielding a $2\times$ difference. Our implementation is fully PyTorch-compatible and integrates seamlessly with training frameworks that use volumetric patch-based sampling for tasks such as segmentation, classification, and detection.

\section{Additional visualizations}
\begin{figure*}[t]
  \centering
  \begin{subfigure}{0.93\linewidth}
    \includegraphics[width=1.0\linewidth]{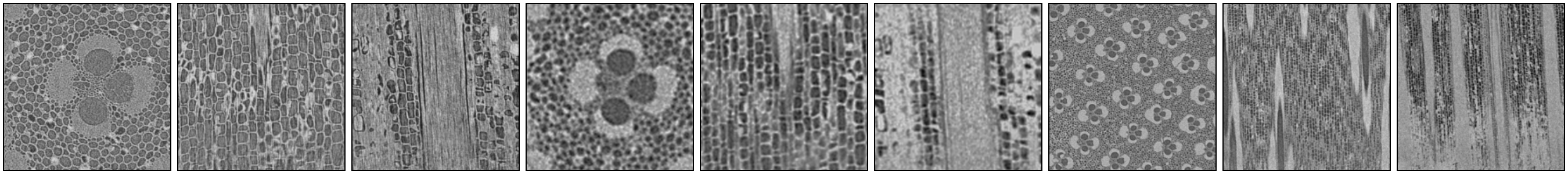}
  \end{subfigure}
  \hfill
  \begin{subfigure}{0.93\linewidth}
    \includegraphics[width=1.0\linewidth]{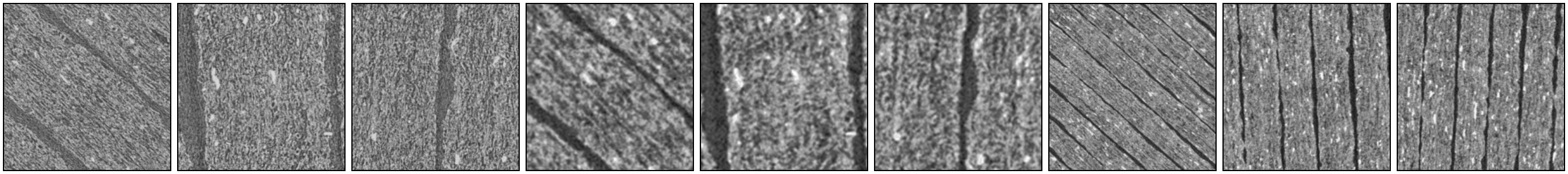}
  \end{subfigure}
  \hfill
  \begin{subfigure}{0.93\linewidth}
    \includegraphics[width=1.0\linewidth]{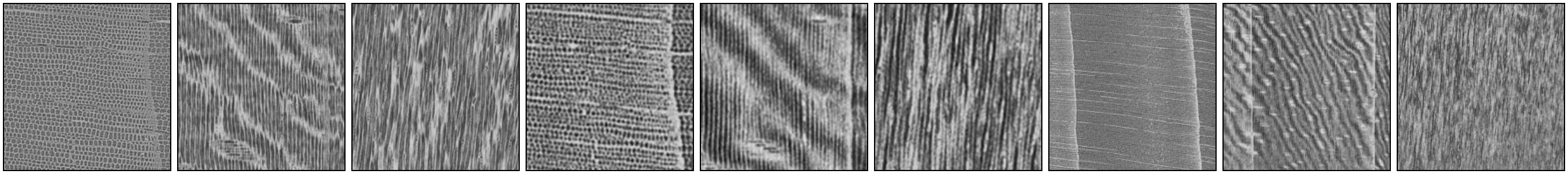}
  \end{subfigure}
  \hfill
  \begin{subfigure}{0.93\linewidth}
    \includegraphics[width=1.0\linewidth]{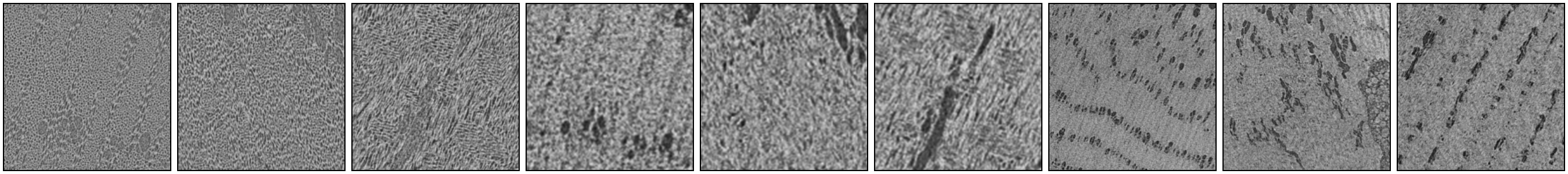}
  \end{subfigure}
  \hfill
  \begin{subfigure}{0.93\linewidth}
    \includegraphics[width=1.0\linewidth]{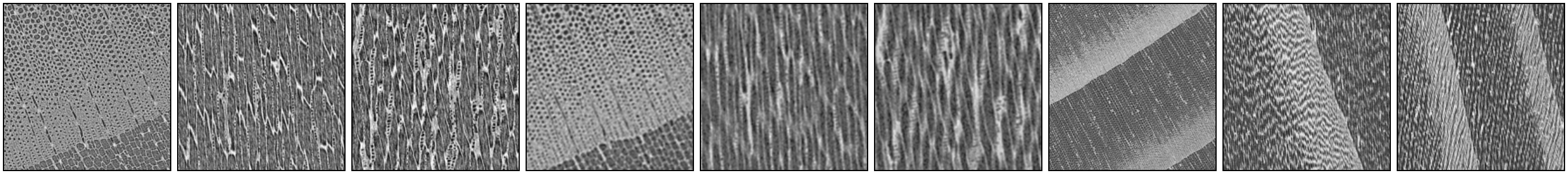}
  \end{subfigure}
  \hfill
  \begin{subfigure}{0.93\linewidth}
    \includegraphics[width=1.0\linewidth]{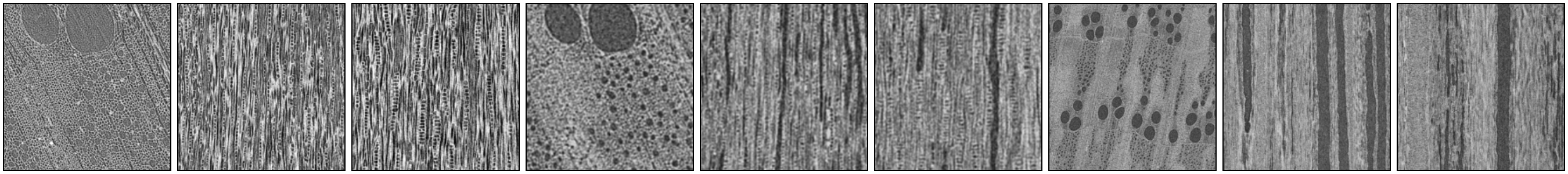}
  \end{subfigure}
  \hfill
  \begin{subfigure}{0.93\linewidth}
    \includegraphics[width=1.0\linewidth]{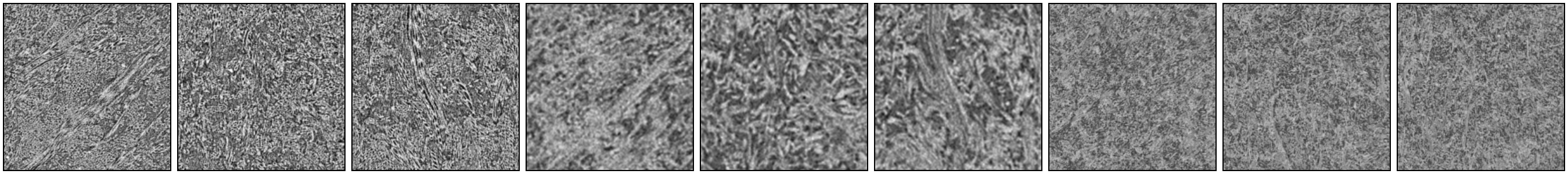}
  \end{subfigure}
  \hfill
  \begin{subfigure}{0.93\linewidth}
    \includegraphics[width=1.0\linewidth]{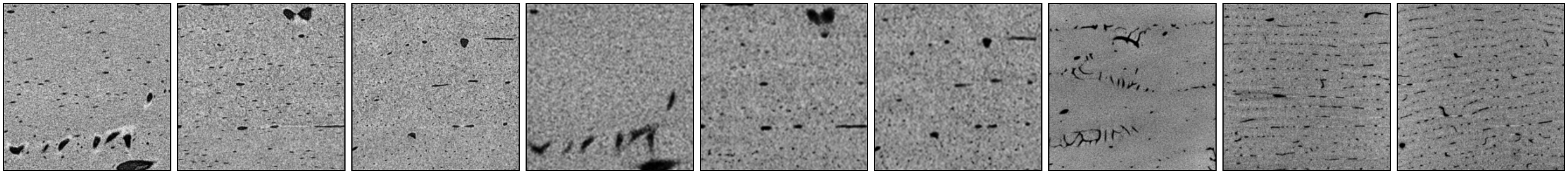}
  \end{subfigure}
  \hfill
  \begin{subfigure}{0.93\linewidth}
    \includegraphics[width=1.0\linewidth]{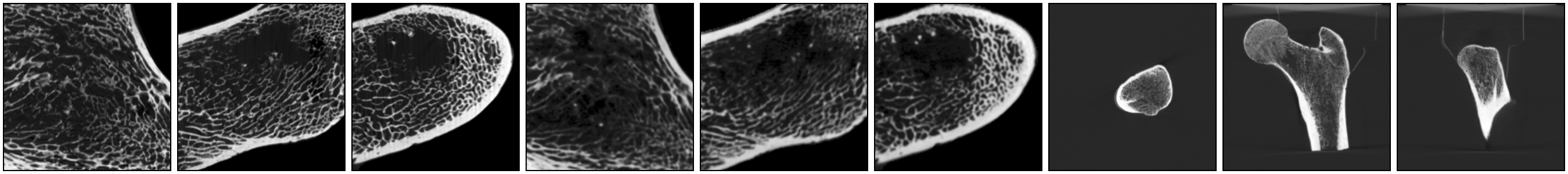}
  \end{subfigure}
  \hfill
  \begin{subfigure}{0.93\linewidth}
    \includegraphics[width=1.0\linewidth]{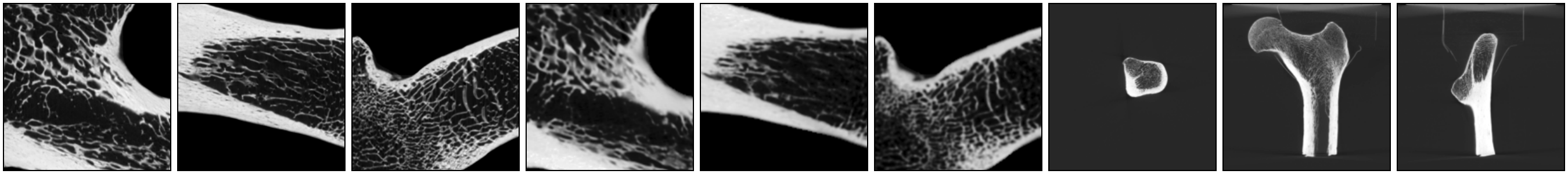}
  \end{subfigure}
  \hfill
  \begin{subfigure}{0.93\linewidth}
    \includegraphics[width=1.0\linewidth]{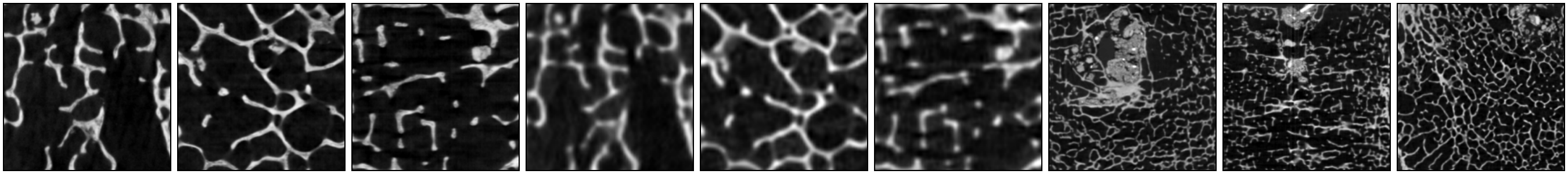}
  \end{subfigure}
  \hfill
  \begin{subfigure}{0.93\linewidth}
    \includegraphics[width=1.0\linewidth]{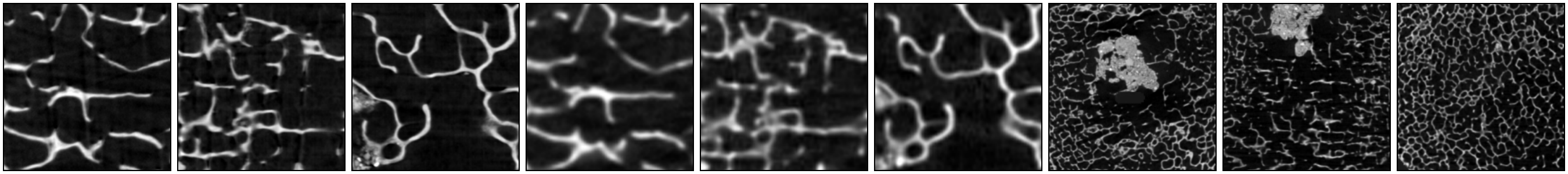}
  \end{subfigure}
  
  \caption{Orthogonal slices from \datasetname, including high-resolution (left), registered (middle) and unregistered LR slices (right). }
  \label{fig:supp_examples}
\end{figure*}

\cref{fig:supp_TV_visualizations} shows additional visualizations of SR model predictions using different training and evaluation data configurations from \datasetname at scale $4\times$. Using downsampled LR data for training but real LR input data for evaluation results in distorted model predictions, highlighting the difference between the two data domains. \cref{fig:supp_examples} provides a showcase of orthogonal image slices from \datasetname, including HR, registered LR and unregistered LR slices. Images are normalized for the purpose of visualization.

\section{Training time}

\cref{tab:supp_training_duration} summarizes the average training time of SR models at scale $4\times$. Training time is measured as the time to complete 100K training iterations averaged across all datasets. 

\section{Evaluation metrics and frequency analysis}

\begin{figure*}[t]
  \centering
  \begin{subfigure}{0.90\linewidth}
    \includegraphics[width=1.0\linewidth]{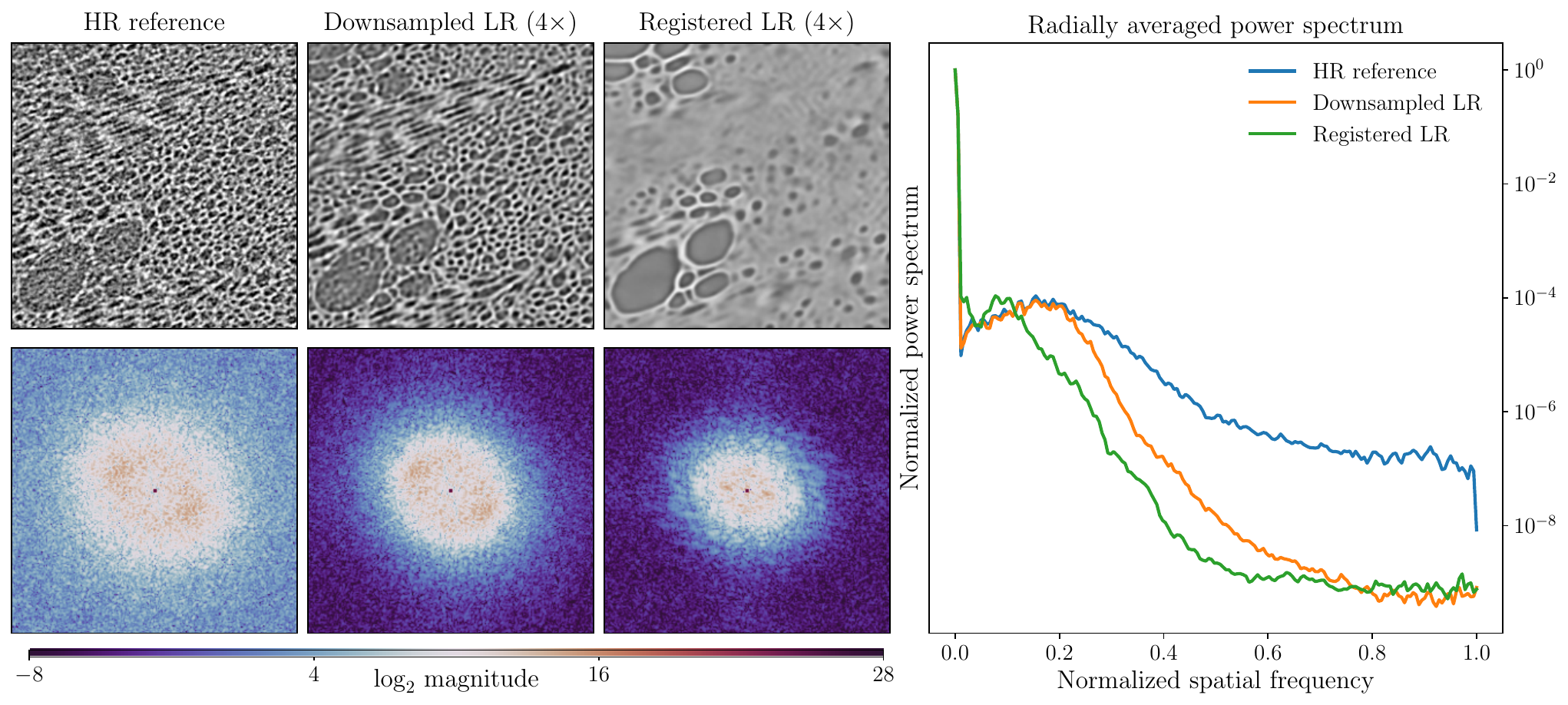}
  \end{subfigure}
  \hfill
  \begin{subfigure}{0.90\linewidth}
    \includegraphics[width=1.0\linewidth]{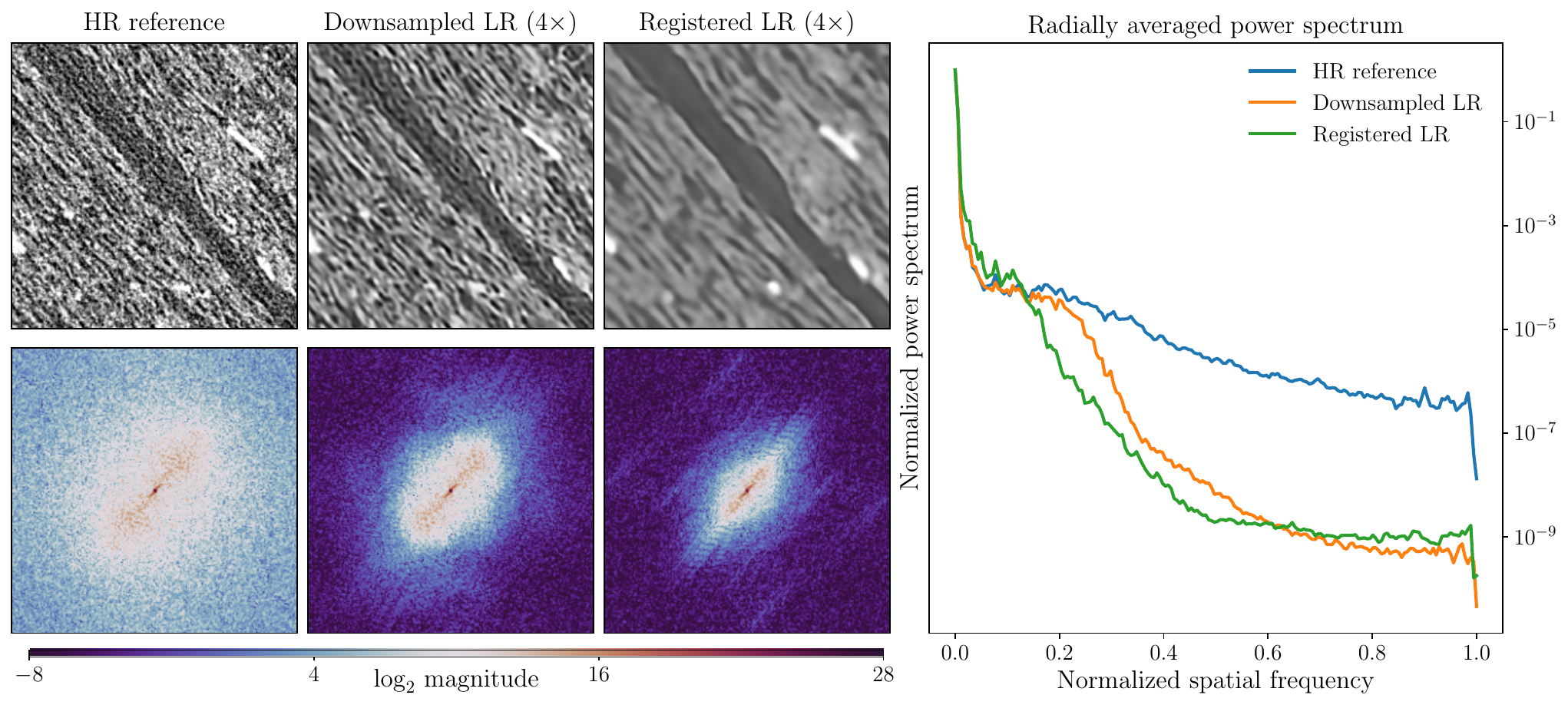}
  \end{subfigure}
  \hfill
  \begin{subfigure}{0.90\linewidth}
    \includegraphics[width=1.0\linewidth]{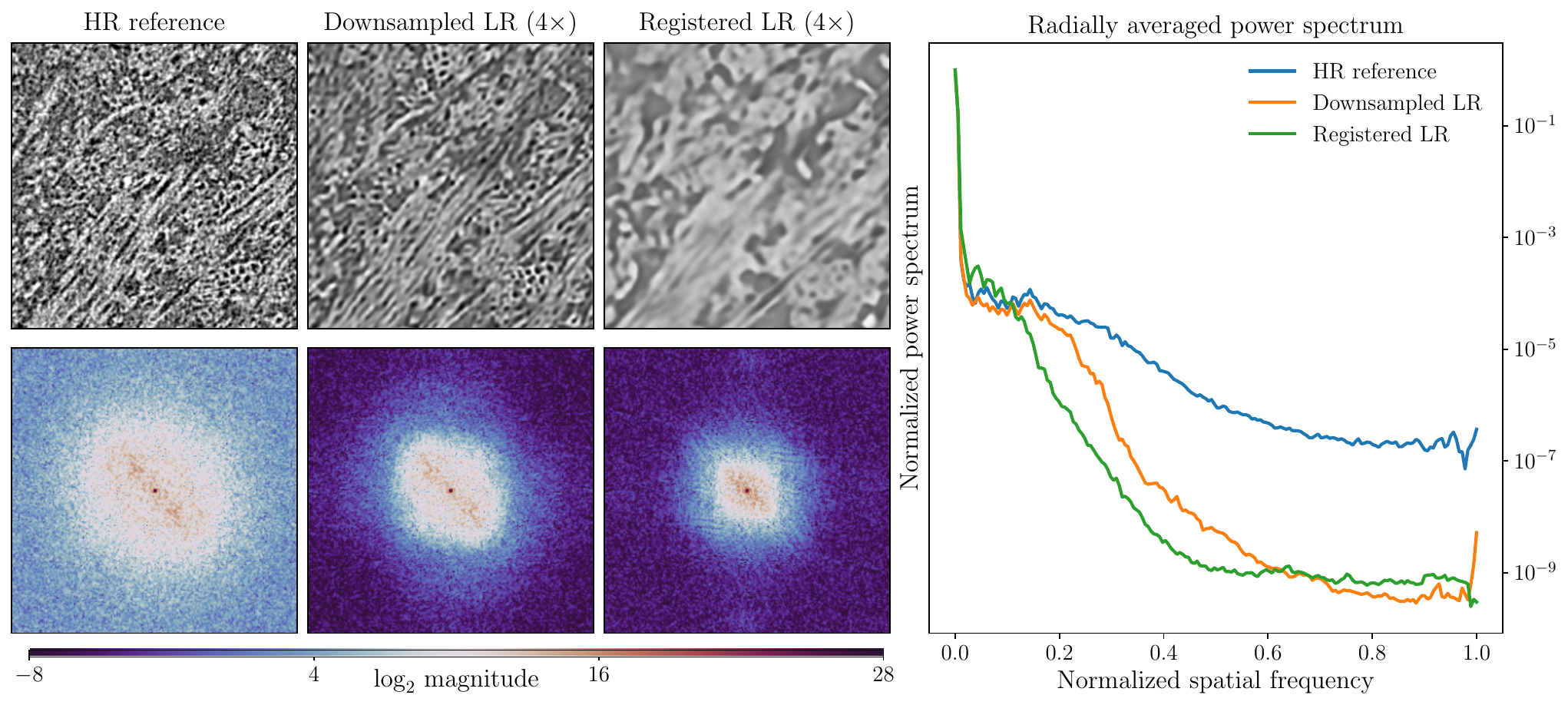}
  \end{subfigure}

  \caption{Comparison of spatial frequency distributions for HR images, and SR predictions using downsampled and real LR images from \datasetname using RRDBNet3D at scale $4\times$. The bottom row shows the log$_2$ power spectra computed from the FFTs of the corresponding images in the top row. Radially averaged power profiles (right) show the relative distribution of power as a function of spatial frequency.}
  \label{fig:supp_power_examples}
\end{figure*}

We report PSNR, SSIM, NRMSE and LPIPS for quantitative evaluation and include total variation (TV) as an indicator of spatial smoothing. While TV captures reductions in local variation in model predictions, it does not distinguish between the removal of noise and the loss of meaningful high-frequency structure. Therefore, our interpretation of TV is done together with visual inspection, which clearly illustrates the characteristic smoothing effect observed when training on real LR data. To further analyze frequency characteristics, we additionally compute power spectrum visualizations and radial frequency profiles for three slice examples from \datasetname, see \cref{fig:supp_power_examples}. As spatial frequency increases, we find that SR predictions derived from scanned LR data exhibit faster decline in signal power compared with SR predictions derived from downsampled images.

\end{document}